\documentclass[11pt,final]{article}
\usepackage{fullpage}
\usepackage[numbers,compress]{natbib}
\newcommand{\citeauth}[1]{\citet{#1}}

\usepackage{times}  
\usepackage{helvet}  
\usepackage{courier}  
\usepackage[hyphens]{url}  
\usepackage{graphicx} 
\usepackage{caption} 

\usepackage{algorithm}
\usepackage{algorithmic}

\usepackage{hyperref}       
\hypersetup{
     colorlinks   = true,
	 citecolor = teal,
}

\usepackage{newfloat}
\usepackage{listings}

\newcount\Includeappendix 
\newcount\preparedraft 
\newcommand{\MDPPF}{\small\textnormal{\textsf{DFAR}}} 

\newcommand{\STUM}{\small\textnormal{\textsf{STAR}}} 
\newcommand{\OCBA}{\small\textnormal{\textsf{OCBA}}}

\newcommand{\Sub}{\alpha}

\newcommand{\STUMNAME}{\STUM}
\newcommand{\STUMr}{\hyperref[alg: general stochastic tree UP-MDP]{\STUMNAME}}

\newcommand{\MDPPFNAME}{\MDPPF}
\newcommand{\MDPPFr}{\hyperref[alg: PARETO_FRONTIER]{\MDPPFNAME}}

\usepackage{amsmath}
\usepackage{amsthm}
\usepackage{mathtools}

\newtheorem{theorem}{Theorem}
\newtheorem{lemma}{Lemma}

\newtheorem{definition}{Definition}
\newtheorem{proposition}{Proposition}

\newtheorem{corollary}{Corollary}
\newtheorem{example}{Example}

\newtheorem{observation}{Observation}

\newcommand{\model}{\textnormal{PARS-MDP}}

\newcommand\norm[1]{\lVert#1\rVert}
\DeclarePairedDelimiter\abs{\lvert}{\rvert}

\usepackage{cleveref}
\newenvironment{proofof}[1]{\begin{proof}[\textnormal{\textbf{Proof of \Cref{#1}}}]}{\end{proof}} 

\newcommand\ceil[1]{\left\lceil#1\right\rceil}
\newcount\Comments  
\usepackage{pgfplots}
\usepackage{nicefrac}
\usepackage{enumitem}
\usepackage{amssymb}
\usepackage{caption}
\usepackage{subcaption}
\usepackage{tikz}

\usetikzlibrary{automata, arrows, positioning}

\DeclareMathOperator*{\argmax}{arg\,max}

\title{Principal-Agent Reward Shaping in MDPs}
\author{
Omer Ben{-}Porat%
\thanks{%
    {Technion---Israel Institute of Technology (\url{omerbp@technion.ac.il})}}
\and Yishay Mansour%
\thanks{%
    {Tel Aviv University and Google Research (\url{mansour.yishay@gmail.com})}}
\and Michal Moshkovitz%
\thanks{%
    {Bosch Center for Artificial Intelligence (\url{michal.moshkovitz@mail.huji.ac.il})}}
\and Boaz Taitler%
\thanks{%
    {Technion---Israel Institute of Technology (\url{boaztaitler@campus.technion.ac.il}), corresponding author}}
}

\begin{document}

\maketitle

\begin{abstract}
Principal-agent problems arise when one party acts on behalf of another, leading to conflicts of interest. The economic literature has extensively studied principal-agent problems, and recent work has extended this to more complex scenarios such as Markov Decision Processes (MDPs). In this paper, we further explore this line of research by investigating how reward shaping under budget constraints can improve the principal's utility. We study a two-player Stackelberg game where the principal and the agent have different reward functions, and the agent chooses an MDP policy for both players. The principal offers an additional reward to the agent, and the agent picks their policy selfishly to maximize their reward, which is the sum of the original and the offered reward. Our results establish the NP-hardness of the problem and offer polynomial approximation algorithms for two classes of instances: Stochastic trees and deterministic decision processes with a finite horizon.
\end{abstract}

\section{Introduction}
The situation in which one party makes decisions on behalf of another party is common. For instance, in the context of investment management, an investor may hire a portfolio manager to manage their investment portfolio with the objective of maximizing returns. However, the portfolio manager may also have their own preferences or incentives, such as seeking to minimize risk or maximizing their own compensation, which may not align with the investor's goals. This conflict is a classic example of principal-agent problems, extensively investigated by economists since the 1970s (see, e.g., \cite{holmstrom1979moral, laffont2003principal}. The fundamental question in this line of work is how the principal should act to mitigate incentive misalignment and achieve their objectives~\cite{dynamic_principal_agent_hidden_info, optimal_coordination_principal-agent_problems, principal_agent_problem_book, zhuang2020consequences, hadfield2019incomplete, xiao2020optimal, ho2014adaptive}.

While the literature on principal-agent problems is vast, modern applications present new challenges in which recommendation systems are principals and their users are agents. To illustrate, consider a navigation app like Waze. While the app's primary function is to provide users with the fastest route to their destination, it also has internal objectives that may not align with those of its users. For example, the app may incentivize users to explore less frequently used roads or to drive near locations that have paid for advertising. Additionally, the app relies on user reports to identify incidents on roads, but users can be reluctant to report. In this scenario, the navigation app is the principal, and it can incentivize users to act in ways that align with its objectives through gamification or by offering coupons for advertisers' stores, among other strategies. The transition of the agent in the space and therefore the principal-agent interaction can be modeled as a Markov decision process (MDP). While some works consider principal-agent problems over MDPs \cite{zhang2008value, yu2022environment, efficient_plan_participate_constraint, plan_participate_constraint}, the setting remains under-explored. The fundamental question of mitigating misalignment in MDP environments warrants additional research.

In this paper, we contribute to the study of this challenging setting. Specifically, we model this setting as a Stackelberg game between two players, Principal and Agent,\footnote{For ease of exposition, third-person singular pronouns are ``she'' for Principal and ``he'' for Agent.} over a joint MDP. Principal and Agent each have a unique reward function, denoted $R^P$ and $R^A$, respectively, which maps states and actions to instantaneous rewards. Principal receives rewards based on Agent's decision-making policy, and thus she seeks to incentivize Agent using a \textit{bonus reward function} we denote $R^B$. We assume that Principal has a limited budget, modeled as a constraint on the norm of $R^B$. Agent is self-interested and seeks a policy that maximizes his own utility, which is the sum of his reward function $R^A$ and the bonus reward function $R^B$ offered by Principal. By offering bonus rewards to Agent, Principal motivates Agent to adopt a policy that aligns better with her (Principal's) objectives. The technical question we ask is \textit{how should Principal structure the bonus function $R^B$ to maximize her own utility given budget constraints?}

\paragraph{Our Contribution} 
This paper is the first to propose efficient, near-optimal algorithms for the principal-agent reward shaping problem. This problem is considered in prior works and has shown to be NP-hard (see elaborated discussion in Subsection~\ref{subsec:related}).  Prior works therefore propose solutions ranging from mixed integer linear programming to differentiable heuristics. In contrast, in this paper, we identify two broad classes of instances that, despite also being NP-hard, can be approximated efficiently.

The first class is MDPs with a tree layout, which we term \emph{stochastic trees}. In stochastic trees, every state has exactly one parent state that leads to it, and several states can have the same parent. We emphasize that the dynamics are not deterministic, and given a state and an action, the next state is a distribution over the children of the state. Stochastic trees are well-suited for addressing real-world scenarios that involve sequential dependencies or hierarchical decision-making. For example, in supply chain management, upstream decisions like raw material procurement have cascading effects on downstream activities such as manufacturing and distribution. Likewise, in natural resource allocation, the initial extraction or harvesting choices create a pathway of subsequent decisions. 

We devise Stochastic Trees principal-Agent Reward shaping algorithm ($\STUMr$), a fully polynomial-time approximation scheme. It uses a surprising \textit{indifference} observation: Imagine two scenarios, one in which Principal grants no bonus, and another where Principal provides an \emph{efficient} bonus, for a definition of efficiency that we describe in Subsection~\ref{subsec:impamentable}. In both cases, Agent gets the same utility if he best responds. This allows us to adopt a bottom-up dynamic programming approach (see Observation~\ref{lemma: same agent reward}) and show that
\begin{theorem}[Informal statement of Theorem~\ref{thm: general stochastic tree UP-MDP}]
Let the underlying MDP be a $k$-ary tree of depth $H$ and let 
$V^P_\star$ be the optimal utility of Principal's problem with budget $B$. Given any small positive constant $\alpha$, our algorithm $\STUMr$ guarantees a utility of at least $V^P_\star$ by using a budget of $B(1+\alpha)$ and its runtime is~$O\left(|A||S|k(\frac{\abs{S}}{\alpha})^3\right)$. 
\end{theorem}

The second class of problems is \emph{finite-horizon deterministic decision processes} (DDP) \cite{castro2020scalable,post2015simplex}. Unlike stochastic trees, where uncertainty plays a central role, DDPs involve scenarios characterized by a clear cause-and-effect relationship between actions and outcomes. DDPs are suitable for many real-world applications, e.g., robotics and control systems that rely on planning. Importantly, the machinery we develop for stochastic trees fails here. We propose another technique that is based on approximating the Pareto frontier of all utility vectors. We devise 
the Deterministic Finite horizon principal-Agent Reward shaping algorithm~($\MDPPFr$), and prove that
\begin{theorem}[Informal statement of Theorem~\ref{thm:alg_deterministic_dag_UP_MDP_approximation}]
Let the underlying MDP be a DDP with horizon $H$, and let $\varepsilon$ be a small positive constant, $\varepsilon > 0$.
Our algorithm $\MDPPFr$ has the following guarantees:
\begin{itemize}[align=left]
\item[$\bullet$ ($\varepsilon$-discrete rewards)] If the reward functions of Principal and Agent are multiples of $\varepsilon$, $\MDPPFr$ outputs an optimal solution. 
\item[$\bullet$ (general rewards)] For general reward functions, $\MDPPFr$ requires $B+H\varepsilon$ budget to guarantee an additive $H\varepsilon$ approximation of Principal's best utility with budget $B$.
\item[$\bullet$ (runtime)] In both cases, executing $\MDPPFr$ takes runtime of $O(\nicefrac{|S||A|{H^2}}{\varepsilon}\log(\nicefrac{|A|{H}}{\varepsilon}))$. 
\end{itemize}
\end{theorem}

\subsection{Related Work}\label{subsec:related}
Most related to this work are works on environment design and policy teaching~\cite{zhang2008value, yu2022environment,zhang2009policy}. These works address scenarios in which the principal can incentivize the agent through an external budget and adopt the same model as we do. \citeauth{zhang2008value} propose a mixed integer linear programming to tackle this problem, while~\citeauth{yu2022environment} modify the agent to have bounded rationality, thereby obtaining a continuous optimization problem. Additionally, \citeauth{zhang2009policy} employ the same model and study how to implement a predefined policy of the principal. Crucially, none of these works offers efficient approaches with provable approximation guarantees. In contrast, our approach targets instances where we can develop polynomial-time approximation algorithms. 

More recent works on principal-agent interactions over MDPs include those of \citet{plan_participate_constraint, efficient_plan_participate_constraint}. They consider MDPs where the principal chooses the policy for both parties, but the agent can stop the execution at any time. They assume that each party has a different reward function, and thus the principal aims at finding a utility-maximizing policy with the constraint of a positive utility for the agent. In this paper, we assume the agent chooses the policy, and not the principal like in \citeauth{efficient_plan_participate_constraint}. In our work, the power of the principal is to \textit{shape} the agent's reward under a limited budget to improve her utility; thus, the problems and treatment are different.

From a broader perspective, principal-agent problems have received significant attention \cite{laffont2003principal}. By and large, solutions are divided into monetary incentives \cite{dynamic_principal_agent_hidden_info, xiao2020optimal, ho2014adaptive} like contracts \cite{contract_theory,dutting2021complexity} and non-monetary incentives (e.g., Bayesian persuasion \cite{wu2022markov, bayesian_persuasion_information_design}). This work addresses the former, as we assume the principal can provide monetary rewards to the agent. The literature on monetary incentives in such problems addresses complex settings like dynamic interaction \cite{dynamic_principal_agent_hidden_info,battaglini2005long, zhang2021automated} and learning contracts adaptively \cite{ho2014adaptive}, among others. 

Our main optimization problem (that appears in Problem~\ref{eq: principles problem}) work could be cast as both a constrained MDPs problem~\cite{cmdp, cmd_budget_allocation} and a Bi-level optimization problems~\cite{stadie2020learning, wang2022bi, hu2020learning, chen2022adaptive, chen2023learning, chakraborty2023aligning}. In constrained MDP problems, the goal is to find a utility-maximizing policy under a global constraint. In Bi-level optimizations, the problem is typically decomposed into an inner optimization problem, the solution of which becomes a parameter for an outer optimization problem. However, our problem cannot benefit from conventional tools and techniques employed in Bi-level optimizations, as two infinitely close bonus allocations can result in arbitrarily different utilities for the principal.

Reward shaping~\cite{policy_invariance_reward_transformation, methods_advising_rl, dynamic_reward_shaping, reward_shaping_epsisodic_rl}  focuses on modifying the learner's reward function by incorporating domain knowledge. The main goals are accelerating the learning process and guiding the agent's exploration~ \cite{policy_invariance_reward_transformation, randlov1998learning, dynamic_reward_shaping, hu2020learning}. We also aim to shape a reward function but in a way that aligns incentives, which is overlooked in that line of work. Particularly, these works do not consider strategic interaction between two entities as we do. Other related works are papers on poisoning attacks~\cite{banihashem2022admissible, rakhsha2020policy, zhang2009policy}, wherein the designer aims to manipulate the environment to steer a learning agent from his originally optimal policy. Finally, our optimization problem also relates to Stacklberg games~\cite{bacsar1998dynamic}) and inverse reinforcement learning~\cite{survey_irl}.

\section{Model}\label{sec:model}
In this section, we present the model along with several properties. We begin by providing some background and notation on Markov decision processes (MDPs) \cite{Sutton2018,MannorMT-RLbook}. An MDP is a tuple $(S, A, P,R,H)$, where $S$ is the state space, $A$ is the action space where $A(s)\subseteq A$ is the subspace of actions available at state $s$. $P$ is the transition function, $P: S \times A \times S \rightarrow [0,1]$ and $\sum_{s'\in S} P(s,a,s')=1$, which is the probability of reaching a state $s'$ from a state $s$ by acting $a$, and $R$ is the (immediate) reward function, $R:S\times A\ \rightarrow \mathbb{R}$. $H$ is the finite horizon, and we assume that there is  a designated initial state $s_{0}$. A policy $\pi: S \rightarrow A$ is a mapping from states to actions.

Given an MDP $(S,A,P,R,H)$ and a policy $\pi$, we let $V(\pi,S,A,P,R,{H})$ denote the \textit{expected utility} of $\pi$, which is the expected sum of immediate rewards; i.e., $V(\pi,S,A,P,R,{H}) = \mathbb{E}[\sum_{i=0}^{H-1} R(s_i,\pi(s_i))]$, where $s_{i+1} \sim P(s_i, \pi(s_i), \cdot)$. We also let $V_s(\pi,S,A,P,R,{H})$ denote the reward in case we start from any state $s\in S$. For convenience, we denote the set of optimal policies by 
\[
\mathcal{A}(S, A, P, R, H) = \argmax_{\pi}{V(\pi, S, A, P, R, {H})}.
\]
When the objects $(S,A,P, H)$ are known from the context, we drop them and denote $V(\pi,R), V_s(\pi,R)$ and $\mathcal{A}(R)$. Finally, we adopt the $Q$ function (see, e.g., \cite{wiering2012reinforcement}), defined as 
\[
Q^\pi(s,a, R) = R(s,a) + \sum_{s' \in S}{P(s, a, s')V_{s'}(\pi, R)}.
\]
The $Q$ function describes the utility from a state $s$ when choosing action $a$ and playing policy $\pi$ afterward. 

The Principal-Agent Reward Shaping MDPs problem ($\model$)  is a two-player sequential game between players that we term Agent and Principal. Formally, an instance of the $\model$ is a tuple $(S, A, P, R^A, R^P, H, B)$, where $(S, A, P, H)$ are the standard ingredients of MDPs as we explain above. The additional ingredients of our model are the (immediate) reward functions $R^A$ and $R^P$, representing the reward functions of Agent and Principal, respectively. These reward functions are typically different, reflecting the different goals and preferences of the two players. We assume that $R^A$ and $R^P$ are always bounded in the $[0,1]$ interval. The last ingredient of our model is the budget $B$, $B \in \mathbb{R}_+$, which represents the total amount of resources available to Principal to distribute over the state-action space and is determined exogenously. The game is played sequentially:
\begin{enumerate}
    \item Principal picks a \textit{bonus reward function} $R^B$, where  $R^B: S\times A\ \rightarrow \mathbb{R}_+$. Principal is restricted to non-negative bonus functions that satisfy the budget constraint, namely $R^B(s,a) \geq 0$ for every $s\in S, a\in A(s)$ and $\sum_{s,a}{R^B(s, a)} \leq B$. 
    
    \item Agent's strategy space is the set of deterministic policies that map $S$ onto $A$. Agent receives the bonus function $R^B$ on top of his standard reward $R^A$, and picks a policy as the best response to the modified reward $R^A + R^B$.
        
\end{enumerate}

The policy Agent picks as a strategy determines Agent's and Principal's utility. Specifically, Agent's utility is the expected sum of rewards, with respect to $R^A + R^B$, received by following policy $\pi$ in the MDP $(S,A,P,R^A + R^B, H)$, taken over the distribution of states and actions induced by policy $\pi$.\footnote{While the bonus function alters the environment and may prompt Agent to adopt a different optimal policy, it does not guarantee that Agent will realize the full value of the additional utility offered through the bonus. This is akin to \emph{money burning} \cite{hartline2008optimal}. Our machinery is also effective in case Principal's constraint in Problem~\ref{eq: principles problem} is on the \textit{realized} budget. We discuss it further in \ifnum\Includeappendix=1{Appendix~\ref{Principal cost for state-action pair}}\else{the appendix}\fi.} Principal's utility is the expected sum of rewards received by following the same policy $\pi$ selected by Agent in the MDP $(S,A,P,R^P, H)$. Here, the reward function used is Principal's own reward function $R^P$. Given a strategy profile $(\pi, R^B)$, the utilities of Principal and Agent are $V(\pi, R^P)$ and $V(\pi, R^A + R^B)$, respectively. Both players wish to maximize their utilities.

\subsection{$\model$ as an Optimization Problem}\label{subsec: optimization}

We propose describing $\model$ as Principal's optimization problem instead of a game formulation. Intuitively, Agent should best respond to the bonus function by using $\mathcal{A}(R^A + R^B) = \argmax_{\pi}{V(\pi, R^A + R^B)}$. In case several policies maximize ${V(\cdot, R^A + R^B)}$, we break ties by assuming that Agent selects the one that most benefits Principal. Formally, $\pi = \argmax_{\pi \in \mathcal{A}(R^A + R^B)} {V(\pi, R^P)}$. Notice that this is \textit{with} loss of generality, although we discuss a straightforward remedy (see  {\ifnum\Includeappendix=1{Appendix~\ref{section: more then one policy for RB}}\else{the appendix}\fi}). Consequently, Agent's action is predictable, and the main challenge in computing equilibrium strategies is finding Principal's optimal action.

Next, we formulate the model as an optimization problem for Principal. Her goal is to choose the bonus function $R^B$ such that the policy chosen by Agent, $\pi \in \mathcal{A}(R^A + R^B)$, maximizes Principal's utility. Formally,
\begin{align}
& \nonumber \max_{R^B}{V(\pi, R^P)} \\
&  \sum_{s\in S, a\in A}{R^B(s, a)} \leq B \label{eq: principles problem}  \tag{P1}\\ 
& \nonumber  {R^B(s, a) \geq 0} \textnormal{ for every } s\in S, a\in A(s) \\
& \nonumber  \pi \in \mathcal{A}(R^A+R^B)
\end{align}
Clearly, without $R^B$ the optimization can be done efficiently, selecting an optimal policy for Agent that maximizes Principal's expected utility (namely, using the tie-breaking in favor of the Principal). When we introduce the variable $R^B$, the problem becomes computationally hard. Intuitively, we need to select both the bonus rewards $R^B$ and Agent's best response policy $\pi$ simultaneously. This correlation is at the core of the hardness; the following Theorem~\ref{thm: UP-MDP NP Hard knapsack} shows that the problem is NP-hard.

\begin{theorem}\label{thm: UP-MDP NP Hard knapsack}
Problem~\ref{eq: principles problem} is NP-hard.
\end{theorem}
We sketch the proof of Theorem~\ref{thm: UP-MDP NP Hard knapsack} in Example~\ref{example:sketch} below. The proof of this theorem, as well as other missing proofs of our formal statements, appear in {\ifnum\Includeappendix=1{Appendix~\ref{appendix: np-hard}}\else{the appendix}\fi}.

\subsection{Warmup Examples} \label{subsection: warmup example}
To get the reader familiar with our notation and illustrate the setting, we present two examples. 

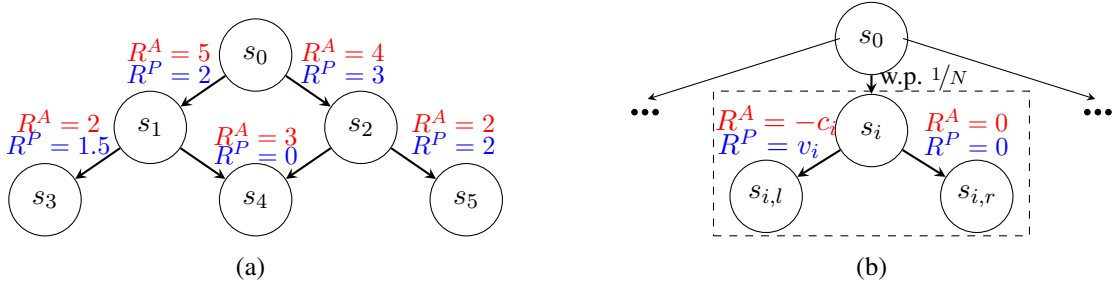
\begin{figure*}
\centering
\begin{subfigure}[b]{.5\textwidth}
  \centering

\begin{tikzpicture}[scale=0.35]
    \node (s0) [state] at (0, 0) {$s_0$};
    \node (s1) [state] at (-4, -2.75) {$s_1$};
    \node (s2) [state] at (4, -2.75) {$s_2$};
    \node (s3) [state] at (-8, -5.5) {$s_3$};
    \node (s4) [state] at (0, -5.5) {$s_4$};
    \node (s5) [state] at (8, -5.5) {$s_5$};

    \node[red,scale=0.9] at (-3.3, 0.2) {$R^A=5$};
    \node[blue,scale=0.9] at (-3.3, -0.65) {$R^P=2$};

    \node[red,scale=0.9] at (3.3, 0.2) {$R^A=4$};
    \node[blue,scale=0.9] at (3.3, -0.65) {$R^P=3$};

    \node[red,scale=0.9] at (-7.5, -2.5) {$R^A=2$};
    \node[blue,scale=0.9] at (-7.5, -3.35) {$R^P=1.5$};

    \node[red,scale=0.9] at (0, -3.0) {$R^A=3$};
    \node[blue,scale=0.9] at (0, -3.7) {$R^P=0$};

    \node[red,scale=0.9] at (7.5, -2.5) {$R^A=2$};
    \node[blue,scale=0.9] at (7.5, -3.35) {$R^P=2$};

    \path[-stealth, thick, sloped, auto]
        (s0) edge (s1)
        (s0) edge (s2)
        (s1) edge (s3)
        (s1) edge (s4)
        (s2) edge (s4)
        (s2) edge (s5);

\end{tikzpicture}

  \caption{}\label{fig:example1}
\end{subfigure}%
\begin{subfigure}[b]{.5\textwidth}
  \centering

    \begin{tikzpicture}[scale=0.35]    
        \node (s0) [state] at (0, 0) {$s_0$};
        \node (si) [state] at (0, -3.5) {$s_i$};
        \node (si1) [state] at (4, -6) {$s_{i,r}$};
        \node (si2) [state] at (-4, -6) {$s_{i,l}$};

        \draw[draw=black, dashed] (-6, -7.5) rectangle (6, -2);

        \draw [-stealth](1.2, 0) -- (8.5, -2.3);
        \draw [-stealth](-1.2, 0) -- (-8.5, -2.3);

        \filldraw (8.2, -2.8) circle (3pt);
        \filldraw (8.6, -2.8) circle (3pt);
        \filldraw (9, -2.8) circle (3pt);

        \filldraw (-8.2, -2.8) circle (3pt);
        \filldraw (-8.6, -2.8) circle (3pt);
        \filldraw (-9, -2.8) circle (3pt);

        \node[scale=0.9] at (0.5, -1.5) {\qquad \quad  w.p. $\nicefrac{1}{N}$};

        \node[red, scale=0.9] at (3.6, -3.1) {$R^A=0$};
        \node[blue, scale=0.9] at (3.6, -4) {$R^P=0$};

        \node[red] at (-3.6, -3.1) {$R^A=-c_i$};
        \node[blue] at (-3.9, -4) {$R^P=v_i$};
        
        \path[-stealth, thick, auto]
            (s0) edge (si)
            (si) edge (si1)
            (si) edge (si2);

    \end{tikzpicture}

  \caption{}\label{fig:example2}%
\end{subfigure}
\caption{Instances for Examples \ref{example:dag} and \ref{example:sketch}. In both figures, Agent's (Principal's) reward is described in red (blue) next to each edge. Figure~\ref{fig:example1} describes an acyclic graph with deterministic transitions, and Figure~\ref{fig:example2} describes a stochastic tree of depth 2. 
}
\label{fig: warmup example}
\end{figure*}

\begin{example}\label{example:dag}
\normalfont
Consider the example illustrated in Figure~\ref{fig:example1}. The underlying MDP has an acyclic \emph{layout},\footnote{We use the term layout to describe the underlying structure of the states, actions, and transition probabilities.} and the transition function is deterministic. The horizon is $H=2$; therefore, the states  $s_3,s_4$ and $s_5$ are \emph{terminal}. At each non-terminal state, Agent chooses action from $\{left, right\}$. The rewards of Agent and Principal are colored (red for Agent, blue for Principal) and appear next to edges, which are pairs of (state, action). In this example, the state-action pairs of $(s_1, right)$ and $(s_2, left)$ share the same rewards and appear once in the figure. Furthermore, assume that the budget is limited to 1; i.e., $B=1$.

Since the transitions are deterministic and so are Agent's policies, each policy corresponds to a path from $s_0$ to a leaf. For instance, the policy that always plays $left$ corresponds to the path $s_0,s_1,s_3$. It is thus convenient to have this equivalence in mind and consider paths instead of policies. The best path for Principal is $\tau^B=(s_0, s_2, s_5)$, with utilities of $V(\tau^B, R^P)=5$ to Principal and $V(\tau^B, R^A)=6$ to Agent. However, Agent has a better path: If he plays $\tau^A=(s_0,s_1,s_4)$, he gets $V(\tau^A, R^A)=8$ while Principal gets $V(\tau^A, R^P)=2$. Indeed, this is Agent's optimal path.

Assume Principal picks $R^B$ such that $R^B(s_3, left)=1$ and $R^B(s,a)=0$ for every $s\in S \setminus \{s_3 \}$ and $a \in A$. This is a valid bonus function since it satisfies the budget constraint. In this case, the path $\tau'=(s_0,s_1,s_3)$ generates Agent's utility of $V(\tau', R^A+R^B)=V(\tau', R^A)+V(\tau', R^B)=7+1=8$. Furthermore, Principal's utility under $\tau'$ is $V(\tau', R^P)=3.5$, which is better than her utility under Agent's default path, $\tau^A$ (recall $V(\tau^A, R^P)=2$). Therefore, since Agent's optimal policies are $\tau^A, \tau'$, our tie-breaking assumption from Subsection~\ref{subsec: optimization} suggests he plays $\tau'$. In fact, the above $R^B$ is the optimal solution to Principal's problem in Problem~\eqref{eq: principles problem}.
\end{example}

\begin{example}\label{example:sketch}
\normalfont
Consider the example illustrated in Figure~\ref{fig:example2}. From the initial state $s_0$, the system transitions uniformly at random to one of $N$ gadgets and reaches the state $s_i$ with probability $\nicefrac{1}{N}$ for any $N\in \mathbb N$. In the gadget associated with $s_i$, Agent can choose deterministically whether to transition to the leaf state $s_{i,r}$ or $s_{i,l}$ by choosing $right$ or choosing $left$. The former results in a reward of zero for both players, while the latter yields a negative reward $-c_i$ for Agent and a positive reward $v_i$ for Principal. Since all $left$ actions induce a negative reward to Agent, his default is always to play $right$. To incentivize Agent to play $left$ at $s_i$, Principal has to allocate her budget such that $R^B(s_i,left)$ is greater or equal to the loss Agent incurs, $c_i$. Hence, setting $R^B(s_i,left)=c_i$ suffices. Whenever $B< \sum_{i=1}^N c_i$, Principal must carefully decide which gadgets to allocate her budget. This example is a reduction from the Knapsack problem (see the proof of Theorem~\ref{thm: UP-MDP NP Hard knapsack}).
\end{example}

\subsection{Implementable Policies}\label{subsec:impamentable}
The following definition captures the set of feasible policies, namely policies that Principal can induce by picking a feasible bonus function.
\begin{definition}[$B$-implementable policy]\label{def:implementable_policy}
A policy $\pi$ is $B$-implementable if there exists bonus function $R^B$ such that $\sum_{s,a}{R^B(s,a)} \leq B$ and $\pi \in  \mathcal{A}(R^A+R^B)$.
\end{definition}
For instance, in Example~\ref{example:dag}, a policy that induces the path $\tau'=(s_0,s_1,s_3)$ is $1$-implementable but not $\nicefrac{1}{2}$-implementable. We further highlight the minimal implementation of a policy.
\begin{definition}[Minimal implementation] \label{def: minimal implementation}
Let $\pi$ be any $B$-implementable policy. We say that $R^B$ is the \textit{minimal implementation} of $\pi$ if $\pi\in \mathcal A(R^A+R^B)$ and for every other bonus function $R$ such that $\pi\in \mathcal A(R^A+R)$, it holds that $\sum_{s,a}{R^B(s,a)} \leq \sum_{s,a}{R(s,a)}$.
\end{definition}
Minimal implementations are budget efficient and refer to the minimal bonus that still incentivizes Agent to play  $\pi$. Importantly, as we prove in {\ifnum\Includeappendix=1{Appendix~\ref{appn: model definition}}\else{the appendix}\fi}, they always exist.

\section{Stochastic Trees} \label{sec: stochastic trees}

In this section, we focus on instances of $\model$ that have a tree layout. To approximate the optimal reward, we propose the Stochastic Trees principal-Agent Reward shaping algorithm ($\STUMr$), which guarantees the optimal Principal utility assuming a budget of $B+\varepsilon  \abs{S}$. Here, $B$ represents the original budget, $ \abs{S}$ is the number of states, and $\varepsilon$ is a configurable discretization factor. Notably, $\STUMr$ has a runtime complexity of $O(|A||S|k(\nicefrac{B}{\varepsilon})^3)$. Before proceeding, we assert that tree-based instances can still be computationally challenging even though they are a special case of the problem. To see this, recall that we used the shallow tree in Example~\ref{example:sketch} to prove the problem is NP-hard. Crucially, it is worth noting that the computational challenge does not stem from the tree structure itself but rather from the presence of randomness. In the case of \emph{deterministic} trees, the trajectory of any policy starting from $s_0$ always leads to a leaf node. Consequently, the Principal's task simplifies to selecting a leaf node from at most $O(|S|)$ leaves.

To begin, we introduce some useful notations. We adopt the standard tree graph terminology of children and parents, where if there is $a \in A$ such that $P(s, a, s') > 0$, then $s$ is a parent of $s'$, and $s'$ is a child of $s$. For convenience, we define the set of children and parent states of state $s$ as $Child(s)$ and $Parent(s)$, respectively. Since the depth of the tree is bounded by the horizon, we can use $H$ to denote an upper bound on the tree's depth, which is defined as the longest path from state $s_{0}$ to a leaf state. Additionally, we denote (any arbitrary)  optimal policy of Agent for $B=0$, which we denote as $\pi^A\in\mathcal{A}(R^A)$. This policy represents the default actions of Agent when there are no bonus rewards.

We present $\STUMr$ formally in the next subsection and sketch the high-level intuition here. It employs a dynamic programming approach, starting from leaf states and iterating toward the root $s_{0}$, while propagating almost all seemingly optimal partial solutions upstream. We use "almost", since it uses a form of discretization (recall that the problem is NP-hard). To explain why this dynamic programming is non-trivial, fix any arbitrary internal (not a leaf) state $s \in S$.  Assume for the moment that Principal places no bonuses; hence, Agent chooses the action in $s$ according to $\pi^A(s)$. Due to $\pi^A$'s optimality w.r.t. $R^A$, we know that $\pi^A(s) \in \argmax_{a\in A(s)} Q^{\pi^A}(s, a, R^A)$. To incentivize Agent to select an action $a' \in A(s), a' \neq \pi^A(s)$, Principal can allocate a bonus according to
\begin{align}\label{eq:diff for minimal}
& R^B(s, a') = Q^{\pi^A}(s, \pi^A(s), R^A) - Q^{\pi^A}(s, a', R^A). 
\end{align}
The term on the right-hand side of Equation~\eqref{eq:diff for minimal}, conventionally referred to as the \emph{advantage function} and represented with a minus sign, plays a significant role in reinforcement learning (see, e.g., \cite{wiering2012reinforcement}).
Note that this bonus includes only the (instantaneous) reward for the pair $(s,a')$. However, any dynamic programming procedure propagates partial bonus allocations; thus, when converting Agent from $\pi^A(s)$ to $a'$, we must consider the bonus allocation in $s$'s subtree. That is, to set $R^B(s, a')$ for converting Agent to playing $a'$ at $s$, we should consider not only $R^A$ but also any candidate bonus function $R^B$ we propagate, and consider Agent's best response. This can result in a significant runtime blowup and budget waste due to discretization.

Fortunately, we can avoid this blowup. The next Observation~\ref{lemma: same agent reward} asserts that, under minimal implementation bonus functions, converting Agent to play another action in $s$ at any state $s$ is decoupled from the allocation at $s$'s subtree.
\begin{observation} \label{lemma: same agent reward}
Let $\pi$ be any $B$-implementable policy, and let $R^B$ be its minimal implementation. For every $s \in S$, it holds that $V_s(\pi^A, R^A) = V_s(\pi, R^A + R^B).$
\end{observation}
Observation~\ref{lemma: same agent reward} is non-intuitive at first glance. On the left-hand side, we have $V_s(\pi^A, R^A)$, which is what Agent gets in the absence of Principal and bonus rewards. On the right-hand side, $V_s(\pi, R^A + R^B)$ is the optimal utility of Agent when Principal picks $R^B$, which is a minimal implementation of the policy $\pi$. Equating the two terms suggests that by granting the bonus reward, Principal makes Agent indifferent between his default policy $\pi^A$ and the $B$-implementable one; namely, the Agent's utility does not increase after placing the bonus reward. 
This is true recursively throughout the tree. Consequently, as long as we consider minimal implementations, we can use the instantaneous bonus proposed in Equation~\eqref{eq:diff for minimal} to incentivize Agent to play $a'$ at $s$ regardless of the bonus allocation in $s$'s subtree.
\begin{algorithm}[t]
\textbf{Input:} $S,A,P,R^P,R^A, H, B, \varepsilon$ \\
\textbf{Output:} $R^B$
\begin{algorithmic}[1]
\small
\caption{Stochastic Trees principal-Agent Reward shaping ($\STUM$)} \label{alg: general stochastic tree UP-MDP}
\STATE let $\mathcal{B} = \{0, \varepsilon, 2\varepsilon, ... , B \}$\label{STUM line: init B}
\STATE for every $s\in S$, $a \in A$, and $b \in \mathcal{B}$, set $U^P(s, b) \gets 0$ and $U^P_a(s, b) \gets 0$ \label{STUM: initialize Up and Upa}
\STATE $curr \gets Leaves(S)$ \label{STUM: initialize curr}
\WHILE{$curr \neq \emptyset$} \label{STUM: curr iteration}
    \STATE pop $s \gets curr$ with the highest depth \label{STUM: pop from curr}

    \STATE for every $a \in A(s)$, set $r(a) \gets Q^{\pi^A}(s, \pi^A(s), R^A) - Q^{\pi^A}(s, a, R^A)$ \label{STUM line: update r of a} 

    \STATE for every $a \in A(s)$ and $b \in \mathcal{B}$, set $U^P_a(s, b) \gets R^P(s, a) + \OCBA(s,a,b)$ \label{STUM line: update T of probability}

    \STATE for every $b\in \mathcal{B}$, set $U^P(s, b) \gets \underset{a \in A(s)}{\max} \{U^P_a(s, b - \underset{\substack{r \in \mathcal{B} \\ r \leq r(a)}}{\max}\{r\})\}$ \label{STUM line: update T of action}

    \STATE $curr \gets curr \cup Parent(s)$ \label{STUM: curr add parent}
\ENDWHILE

\STATE extract $R^B$ from $U^P$ and $U^P_a$ \label{STUM line: extract RB from TK}
\RETURN $R^B$

\end{algorithmic}
\end{algorithm}

\subsection{The $\STUM$ Algorithm}
$\STUMr$ is implemented in Algorithm~\ref{alg: general stochastic tree UP-MDP}. It gets the instance parameters as input, along with a discretization factor $\varepsilon$, and outputs an almost optimal bonus function $R^B$. Line~\ref{STUM line: init B} initializes the discrete set of bonuses $\mathcal B$, referred to as \textit{budget units}. Line~\ref{STUM: initialize Up and Upa} initializes the variables $U^P$ and $U^P_a$, which we use to store partial optimal solutions for Principal. We then initialize $curr$ to  $Leaves(S)$ in Line~\ref{STUM: initialize curr}, where  $Leaves(S)$ is the set of leaf states. 
The backward induction process is the while loop in Lines~\ref{STUM: curr iteration}. Throughout the execution, $curr$ stores the states whose subtrees were already processed in previous iterations. We iterate while $curr$ is non-empty. Line~\ref{STUM: pop from curr}, we pop a state $s$ from $curr$. In Line~\ref{STUM line: update r of a}, we compute for every action $a\in A(s)$ the minimal bonus needed to shift Agent from playing $\pi^A(s)$. According to Observation~\ref{lemma: same agent reward}, this local deviation is decoupled from partial optimal solutions we computed for $Child(s)$ for minimal bonus functions.

Lines~\ref{STUM line: update T of probability} and~\ref{STUM line: update T of action} are the heart of the dynamic programming process. In Line~\ref{STUM line: update T of probability}, we consider every action $a\in A(s)$ and every budget unit $b\in \mathcal{B}$. We set $U_a^P(s,b)$ to be the highest (expected) utility of Principal when starting from $s$, assuming Agent plays $a$ and the budget is $b$. For that, we need to address two terms. The first term is the local $R^P(s,a)$, the reward Principal gets if Agents plays $a$ at $s$. The second term in Line~\ref{STUM line: update T of probability} is $\OCBA$, which stands for Optimal Children Budget Allocation and is based on inductive computation. $\OCBA(s,a,b)$ is Principal's optimal utility if Agent plays action $a$ in state $s$ \textit{and} we allocate  the budget $b$ optimally among the subtrees of $Child(s)$. $\OCBA(s,a,b)$ can be computed in $O(k (\nicefrac{b}{\varepsilon})^2)$ time via another (and different) dynamic programming process. Due to space limitations, we differ the implementation of $\OCBA$  for $k$-ary trees to {\ifnum\Includeappendix=1{Appendix~\ref{apn subsection: budget to k children}}\else{the appendix}\fi} and explain for binary trees. Let $s_r, s_l$ be the right and left child of $s$, respectively. Then, $\OCBA(s,a,b)$ is 
{\small
\begin{equation*}
\max_{\substack{b' \in \mathcal{B},b' \leq b}} \left\{ P(s, a, s_r)U^P(s_r, b') + 
    P(s, a, s_l)U^P(s_l, b - b')
 \right\}.    
\end{equation*}
}

Namely, we consider all possible budget allocations between the subtrees of the children $s_r$ and $s_l$, allocating $b'$ to the former and $b-b'$ to the latter. 
The second step of the dynamic programming appears in Line~\ref{STUM line: update T of action}, where we compute $U^P(s,b)$ inductively. Recall that the previously computed $U_a^P$ assumed Agent plays $a$ for $a\in A(s)$, but did not consider the required budget for that. To motivate Agent to play $a$, we need to allocate the bonus $r(a)$ to the pair $(s,a)$ (recall Line~\ref{STUM line: update r of a}). Hence, $U^P(s,b)$ is the maximum over $U_a^P$ computed in the previous line, but considering that we must allocate $R^B(s,a)=r(a)$. Since we consider discrete budget, we assume we exhaust the minimal budget unit $r\in \mathcal B$ greater than $r(a)$, meaning we slash a slight budget portion in the inductive process.

After the inductive computation, Line~\ref{STUM: curr add parent} updates the set $curr$. Line~\ref{STUM line: extract RB from TK} extracts the optimal bonus reward $R^B$ from $U^P$ and $U^P_a$. To that end, we use backtracking, which involves tracing the actions that led to the optimal value of $U^P$ while considering the required bonus reward for Agent's deviations. The backtracking identifies the best sequence of actions and corresponding bonus rewards, which we formally claim later are the (approximately) best $B$-implementable policy and its corresponding minimal implementation. We end this subsection with the formal guarantees of $\STUMr$.
\begin{theorem} \label{thm: general stochastic tree UP-MDP}
Let $I = (S, A, P, R^A, R^P, H, B)$ be a $k$-ary tree, and let $V^P_\star$ be the optimal solution for $I$. Further, let $\tilde{I}$ be the identical instance but with a budget $B + \varepsilon \abs{S}$ for a small constant $\varepsilon > 0$, and let $\tilde{R}^{B}$ denote the output of $\STUMr(\tilde I)$. Then,  executing $\STUMr(\tilde I)$ takes a  run time of $O\left(|A||S|k(\nicefrac{B}{\varepsilon})^3\right)$, and its output $\tilde{R}^{B}$ satisfies $V(\pi, R^P) \geq V^P_\star$ for any $\pi \in \mathcal{A}(R^A + \tilde{R}^{B})$.
\end{theorem}

We note that $\STUMr$  is a fully polynomial-time approximation scheme (FPTAS) despite the factor $B$ in the runtime in Theorem~\ref{thm: general stochastic tree UP-MDP}. Let $\alpha > 0$ be any small constant. By setting $\varepsilon=\frac{B \alpha}{ \abs{S}}$, we use a budget of $B(1+\alpha)$ and the execution takes $O\left(|A||S|k(\frac{ \abs{S}}{\alpha})^3\right)$.

\section{Deterministic Decision Processes with Finite Horizon} \label{sec: deterministic DAG}

This section addresses $\model$ instances with a deterministic decision process layout and finite horizon.  As we show in {\ifnum\Includeappendix=1{Appendix~\ref{sec: deterministic hard}}\else{the appendix}\fi}, this class of problems is still NP-hard. We propose the Deterministic Finite horizon principal-Agent Reward shaping algorithm ($\MDPPFr$), which is implemented in Algorithm~\ref{alg: PARETO_FRONTIER}. In case  $R^A$ and $R^P$ are \emph{$\varepsilon$-discrete}, i.e., multiples of some small constant $\varepsilon>0$, $\MDPPFr$ provides an optimal solution to Problem~\eqref{eq: principles problem} and runs in $O(\nicefrac{|S||A|{H^2}}{\varepsilon}\log(\nicefrac{|A|{H}}{\varepsilon}))$ time, where $H$ is the horizon. As a corollary, we show that $\MDPPFr$ provides a bi-criteria approximation for general reward functions $R^A$ and $R^P$.  Namely, if the optimal solution for a budget $B$ is $V^P_{\star}$, $\MDPPFr$ requires a budget of $B+H\varepsilon$ to guarantee a utility of at least $V^P_{\star}-H\varepsilon$.

For ease of readability, we limit our attention to acyclic DDPs and explain how to extend our results to cyclic graphs later in Subsection~\ref{subsec:cyclic}. Before presenting an approximation algorithm for this class of instances, we note that the $\STUMr$ algorithm from the previous section is inappropriate for instances with an acyclic layout. One of the primary challenges in $\STUMr$ is allocating the budget between children states for each action. In an acyclic layout, a state may have multiple parent states; therefore, if we follow the same technique for non-tree layouts, the $\STUMr$ algorithm would assign multiple different budgets to the same state, one budget from each parent. As a result, a single state-action pair may have more than one bonus reward assigned to it. To handle acyclic layouts, we employ different techniques. To explain the intuition behind our algorithm, consider the set of all utility vectors $\mathcal U$, where $\Pi$ is the set of all policies and 
$\mathcal{U} = \left\{ \left(  V(\pi, R^A), V(\pi, R^P) \right) \in \mathbb{R}^2 \mid \pi \in \Pi \right\}$.

Every element in $\mathcal U$ is a two-dimensional vector, where the entries are the utilities of Agent and Principal, respectively. Ideally, we would like to find the best utility vector in $\mathcal U$: One that corresponds to a $B$-implementable policy and maximizes Principal's utility. However, we have two obstacles. First, constructing $\mathcal U$ is infeasible as its size can be exponential. We circumvent this by discretizing the reward functions of both players to be multiples of a small~$\varepsilon$.
Discretizing the reward function ensures that utilities will also be $\varepsilon$-discrete. Since $H$ constitutes an upper bound on the highest utility (recall we assume $R^A,R^P$ are bounded by 1 for every state-action pair), each player can have at most $\nicefrac{H}{\varepsilon}$ different utilities; thus, the $\varepsilon$-discrete set $\mathcal U$ can have at most $(\nicefrac{H}{\varepsilon})^2$ different vectors. The \textit{Pareto frontier}, i.e., the set of Pareto efficient utilities, contains at most $\nicefrac{2H}{\varepsilon}$. Our algorithm propagates the Pareto efficient utilities bottom-up.

The second obstacle is that the Pareto frontier we compute includes utility vectors corresponding to policies that are not $B$-implementable and thus infeasible. The following observation asserts that we can quickly distinguish utility vectors belonging to $B$-implementable policies.
\begin{observation}\label{observation: deterministic and pareto}
If the transition function is deterministic, then a policy $\pi$ is $B$-implementable if and only if $V(\pi, R^A) \geq V(\pi^A, R^A) - B$.
\end{observation}

\subsection{The $\MDPPF$ Algorithm}
\begin{algorithm}[t]
\textbf{Input:} $S,A,P,R^P,R^A, H, B$, where $R^P$ and $R^A$ are assumed to be $\varepsilon$-discrete \\
\textbf{Output:} $R^B$
\begin{algorithmic}[1]
\small 
\caption{Deterministic Finite horizon principal-Agent Reward shaping ($\MDPPF$)} \label{alg: PARETO_FRONTIER}
\STATE for every $s \in S$, set $U(s) \gets \emptyset$ \label{MDPPF init Pareto s}

\STATE for all $s\in Terminal(S)$, let $U(s) \gets \{(0,0) \}$ \label{val_calculation_terminal}
\STATE $S_{pass} \gets Terminal(S)$ \label{MDPPF: initialize curr}
\WHILE {$S_{pass} \neq S$} \label{backward_propagate}
    \STATE select a state $s \in  Terminal(S\setminus S_{pass})$ \label{calc_frontier_for_curr}
    
    \STATE for every $a\in A(s), s'\in Child(s,a)$ and $\mathbf u\in U(s')$, set \label{val_calculation}
    
     \hfill $U(s) \gets U(s) \cup \left( (R^A(s, a), R^P(s, a)) + \mathbf u \right)$\hfill

    \STATE $U(s) \gets Pareto(U(s))$ \label{call_remove_dominated}

    \STATE $S_{pass} \gets S_{pass} \cup \{s\}$ \label{update_curr}
\ENDWHILE
\STATE let $\pi$ such that $u^\pi \gets \underset{\substack{u^\pi \in U(s_{0}) \\ V(\pi, R^A) \geq V(\pi^A, R^A) - B}}{\argmax} V(\pi, R^P)$  \label{deterministic_DAG_search_optimal_val}
\STATE extract $R^B$ from $\pi$ \label{MDPPF: get RB of policy} 
\RETURN{$R^B$} \;
\end{algorithmic}
\end{algorithm}

The $\MDPPFr$ algorithm receives the instance parameters, where we assume $R^A$ and $R^P$ are $\varepsilon$-discrete for some constant $\varepsilon >0$ (Corollary~\ref{corollary:alg_deterministic_dag_UP_MDP_approximation} explains how to relax this assumption), and outputs an \emph{optimal} $R^B$. $\MDPPFr$ begins by initializing the set $U(s)$ to be the empty set for each state $s$ in Line~\ref{MDPPF init Pareto s}. In Line~\ref{val_calculation_terminal}, we set $U(s)$ to include the zero vector for every terminal state. A state is \emph{terminal} if it does not allow transitions to other states, and  $Terminal(S')$ denotes the set of terminal states in the induced graph with states $S' \subseteq S$. Line~\ref{MDPPF: initialize curr} initializes $S_{pass}$ to the set of $Terminal(S)$. Line~\ref{backward_propagate} is a while loop that executes until $S_{pass}$ contains all states.

In Line~\ref{calc_frontier_for_curr}, we pick a state $s$ we have not processed yet, namely, a terminal state of $S\setminus S_{pass}$. Due to the way we process states, $Child(s) \subseteq S_{pass}$. Further, since the graph is acyclic, such a state $s$ must exist. 
In Line~\ref{val_calculation}, we let $Child(s,a)$ denote the state we reach by acting $a\in A(s)$ in $s$ (this state is unique since transitions are deterministic). Due to inductive arguments,  $U(s')$ encompasses all attainable utilities when starting from state $s'$, for every child $s'\in Child(s)$. In other words, every element in $U(s')$ has the form $(V_{s'}(\pi, R^A), V_{s'}(\pi, R^P))$ for some policy $\pi$. We update $U(s)$ to contain all vectors $(R^A(s, a), R^P(s, a)) + \mathbf u $ for $a\in A(s)$ and $\mathbf u \in U(s')$, where $s'= Child(s,a)$. After this update, elements in $U(s)$ represent the $Q$ function vector $(Q^{\pi}(s, a, R^A), Q^{\pi}(s, a, R^P))$ for any $a \in A(s)$.

In Line~\ref{call_remove_dominated}, we remove Pareto inefficient utility vectors. We show in \ifnum\Includeappendix=1{Appendix~\ref{appendix pareto remove complexity} }\else{the appendix }\fi that this can be done in linearithmic time. Finalizing the while loop, Line~\ref{update_curr} updates $S_{pass}$. 
By the time we reach Line~\ref{deterministic_DAG_search_optimal_val}, $U(s_0)$ contains all attainable utility vectors, including infeasible utility vectors that require a budget greater than $B$. Observation~\ref{observation: deterministic and pareto} assists in distinguishing utility vectors associated with $B$-implementable policies. We pick the utility vector $u^\pi$ that maximizes Principal's utility, and work top-down to reconstruct the policy $\pi$ that attains it. Finally, we reconstruct the minimal implementation $R^B$ that induces $\pi$ by setting the right-hand side of Equation~\eqref{eq:diff for minimal} for every pair $(s, \pi(s))$. Next, we present the formal guarantees that $\MDPPFr$ provides.

\begin{theorem}\label{thm:alg_deterministic_dag_UP_MDP_approximation}
Let $I=(S, A, P, R^A, R^P, H, B)$ be an acyclic and deterministic instance, and assume 
the reward function $ R^A$ and $R^P$ are $\varepsilon$-discrete for a small constant $\varepsilon > 0$. Further, let $V^P_\star$ be the optimal solution for $I$. Then,  executing $\MDPPFr(I)$ takes a run time of  $O\left(\frac{|S||A|H}{\varepsilon}\log(\frac{|A|H}{\varepsilon})\right)$, and its output ${R}^B$ satisfies
$V(\pi, R^P) =V^P_\star$ for any $\pi \in \mathcal{A}(R^A +{R}^B)$.
\end{theorem}
We also leverage Theorem~\ref{thm:alg_deterministic_dag_UP_MDP_approximation} to treat reward functions that are not $\varepsilon$-discrete. According to Corollary~\ref{corollary:alg_deterministic_dag_UP_MDP_approximation} below, $\MDPPFr$ provides a bi-criteria approximation for general reward functions $R^A$ and $R^P$. Namely, it requires a budget of $B + H\varepsilon$ to provide an additive approximation of  $H\varepsilon$.

\begin{corollary}\label{corollary:alg_deterministic_dag_UP_MDP_approximation}
Let $I=(S, A, P, R^A, R^P, H, B)$ be an acyclic and deterministic instance, and let $V^P_\star$ be the optimal solution for $I$. Let $\tilde{I} = (S, A, P, \tilde{R}^A, \tilde{R}^P, H, B + H\varepsilon)$ be an instance with $\varepsilon$-discrete versions of $R^A$ and $R^P$, $\tilde{R}^A$ and $\tilde{R}^P$, respectively, for a small constant $\varepsilon$. If $\MDPPFr(\tilde I)$ outputs $\tilde{R}^B$, then it holds that $V(\pi, R^P) \geq  V^P_\star - H\varepsilon$ for any $\pi \in \mathcal{A}(R^A + \tilde{R}^B).$
\end{corollary}

\subsection{Cyclic Deterministic Decision Processes}\label{subsec:cyclic}
We develop $\MDPPFr$ assuming the underlying DDP has no cycles. In this subsection, we address the case of cyclic DDPs. Notice that any DDP with a finite horizon $H$ can be cast as an \emph{acyclic} layer graph with $\abs{S} \cdot H$ states. Due to space limitations, we describe this construction formally in {\ifnum\Includeappendix=1{Appendix~\ref{section: ddp to acyclic layered}}\else{the appendix}\fi}. Crucially, the resulting DDP is acyclic and has $|S|H$ states, compared to $|S|$ states in the original DDP. Consequently, every factor $|S|$ in the runtime guarantees of $\MDPPFr$ should be replaced with $|S|H$; thus, executing $\MDPPFr$ on the modified acyclic DDP instance takes $O(\nicefrac{|S||A|{H^2}}{\varepsilon}\log(\nicefrac{|A|{H}}{\varepsilon}))$.

\section{Simulations}\label{sec:simulations}
\usepgfplotslibrary{fillbetween}
\usepgfplotslibrary{groupplots}

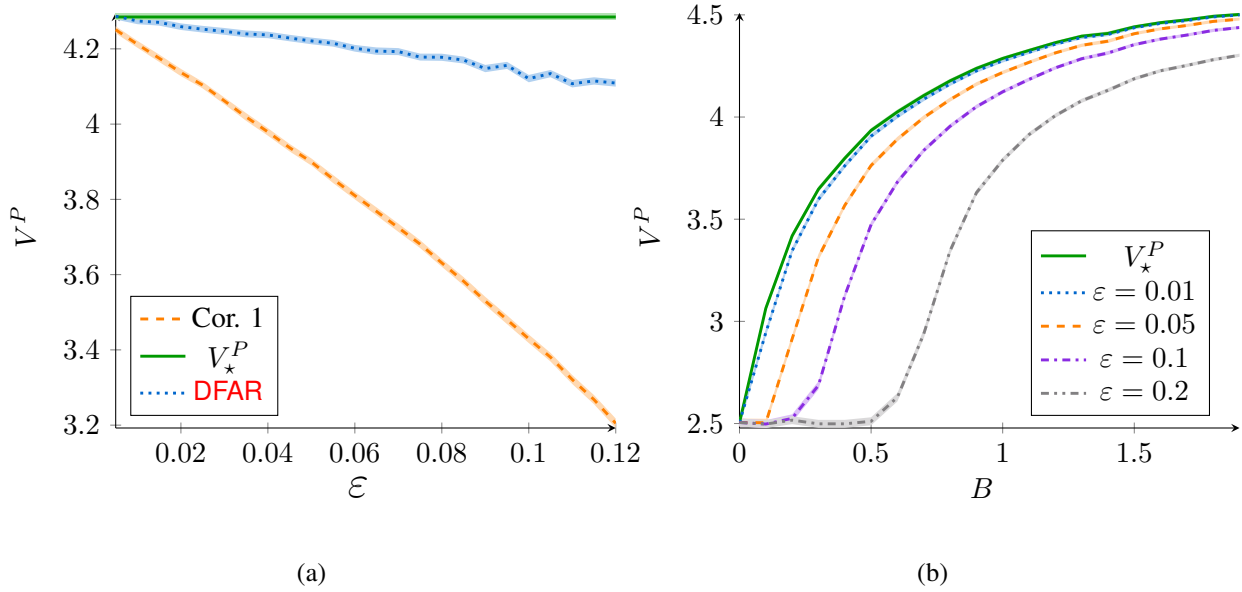
\begin{figure*}[t]
\centering
\begin{subfigure}[b]{.5\textwidth}
  \centering
  \def\datagraphone{simulation/deterministic_graph1.csv}
\definecolor{O1}{RGB}{255,128,0}
\definecolor{O2}{RGB}{0,102,204}
\definecolor{O3}{RGB}{0,153,0}
\definecolor{O4}{RGB}{138,43,226}
\definecolor{O5}{RGB}{255,128,130}
\definecolor{O6}{RGB}{130,128,130}
\def\unicolor{O1}
\def\unione{O2}
\def\unitwo{O3}
\def\unitree{O4}
\def\unifour{O5}
\def\unifive{O6}

\begin{tikzpicture}[scale=0.97]

\begin{axis}[
    axis lines = left,
    xlabel = {$\varepsilon$},
    xlabel style={font=\LARGE},
    xticklabel={
            \pgfmathparse{\tick}\pgfmathprintnumber[fixed, precision=3]{\pgfmathresult}
        },
    ylabel = {\(V^P\)},
    legend pos=south west,
]
\addplot [no markers, \unicolor, line width=1.1pt, dashed] table[x=eps, y=Thm4_mean, col sep=comma] {\datagraphone};
\addlegendentry{\textnormal{Cor. 1}}

\addplot [no markers, \unitwo, line width=1.1pt,solid] table[x=eps, y=opt_mean, col sep=comma] {\datagraphone};
\addlegendentry{$V_\star^P$}

\addplot [no markers, \unione, line width=1.1pt, dotted] table[x=eps, y=DFAR_mean, col sep=comma] {\datagraphone};
\addlegendentry{$\MDPPFr$}

\addplot [name path=unione_upper, draw=none, fill opacity=0.3, forget plot] table[x=eps, y expr=\thisrow{DFAR_mean}+(3*\thisrow{DFAR_var}), col sep=comma] {\datagraphone} ;

\addplot [name path=unione_lower, draw=none, fill opacity=0.3, forget plot] table[x=eps, y expr=\thisrow{DFAR_mean}-(3*\thisrow{DFAR_var}), col sep=comma] {\datagraphone} ;

\addplot [\unione!30, forget plot, fill opacity = 0.5] fill between[of=unione_upper and unione_lower];

\addplot [name path=unicolor_upper, draw=none, fill opacity=0.3, forget plot] table[x=eps, y expr=\thisrow{Thm4_mean}+(3*\thisrow{Thm4_var}), col sep=comma] {\datagraphone} ;

\addplot [name path=unicolor_lower, draw=none, fill opacity=0.3, forget plot] table[x=eps, y expr=\thisrow{Thm4_mean}-(3*\thisrow{Thm4_var}), col sep=comma] {\datagraphone} ;

\addplot [\unicolor!30, forget plot, fill opacity = 0.5] fill between[of=unicolor_upper and unicolor_lower];

\addplot [name path=unitwo_upper, draw=none, fill opacity=0.3, forget plot] table[x=eps, y expr=\thisrow{opt_mean}+(3*\thisrow{opt_var}), col sep=comma] {\datagraphone} ;

\addplot [name path=unitwo_lower, draw=none, fill opacity=0.3, forget plot] table[x=eps, y expr=\thisrow{opt_mean}-(3*\thisrow{opt_var}), col sep=comma] {\datagraphone} ;

\addplot [\unitwo!30, forget plot, fill opacity = 0.5] fill between[of=unitwo_upper and unitwo_lower];

\end{axis}

\end{tikzpicture}
  \caption{}
  \label{simulation fig1}
\end{subfigure}%
\begin{subfigure}[b]{.5\textwidth}
  \centering
  \def\datagraphtwo{simulation/deterministic_graph2.csv}
\definecolor{O1}{RGB}{255,128,0}
\definecolor{O2}{RGB}{0,102,204}
\definecolor{O3}{RGB}{0,153,0}
\definecolor{O4}{RGB}{138,43,226}
\definecolor{O5}{RGB}{255,128,130}
\definecolor{O6}{RGB}{130,128,130}
\def\unicolor{O1}
\def\unione{O2}
\def\unitwo{O3}
\def\unitree{O4}
\def\unifour{O6}
\def\unifive{O6}

\begin{tikzpicture}[scale=0.97]

\begin{axis}[
    axis lines = left,
    xlabel = $B$,
    ylabel = {\(V^P\)},
    legend pos=south east,
]
\addplot [no markers, \unitwo, line width=1.1pt, solid] table[x=B, y=BF_mean, col sep=comma] {\datagraphtwo};
\addlegendentry{$V^P_\star$}

\addplot [no markers, \unione, line width=1.1pt, dotted] table[x=B, y=DFAR_mean_001, col sep=comma] {\datagraphtwo};
\addlegendentry{\(\varepsilon = 0.01\)}

\addplot [no markers, \unicolor, line width=1.1pt, dashed] table[x=B, y=DFAR_mean_005, col sep=comma] {\datagraphtwo};
\addlegendentry{\(\varepsilon = 0.05\)}

\addplot [no markers, \unitree, line width=1.1pt, dashdotted] table[x=B, y=DFAR_mean_01, col sep=comma] {\datagraphtwo};
\addlegendentry{\(\varepsilon = 0.1\)}

\addplot [no markers, \unifour, line width=1.1pt, dashdotdotted] table[x=B, y=DFAR_mean_02, col sep=comma] {\datagraphtwo};
\addlegendentry{\(\varepsilon = 0.2\)}

\addplot [name path=unicolor_upper, draw=none, fill opacity=0.3, forget plot] table[x=B, y expr=\thisrow{BF_mean}+(3*\thisrow{BF_var}), col sep=comma] {\datagraphtwo} ;

\addplot [name path=unicolor_lower, draw=none, fill opacity=0.3, forget plot] table[x=B, y expr=\thisrow{BF_mean}-(3*\thisrow{BF_var}), col sep=comma] {\datagraphtwo} ;

\addplot [\unitwo!30, forget plot, fill opacity = 0.5] fill between[of=unicolor_upper and unicolor_lower];

\addplot [name path=unione_upper, draw=none, fill opacity=0.3, forget plot] table[x=B, y expr=\thisrow{DFAR_mean_001}+(3*\thisrow{DFAR_var_001}), col sep=comma] {\datagraphtwo} ;

\addplot [name path=unione_lower, draw=none, fill opacity=0.3, forget plot] table[x=B, y expr=\thisrow{DFAR_mean_001}-(3*\thisrow{DFAR_var_001}), col sep=comma] {\datagraphtwo} ;

\addplot [\unione!30, forget plot, fill opacity = 0.5] fill between[of=unione_upper and unione_lower];

\addplot [name path=unitwo_upper, draw=none, fill opacity=0.3, forget plot] table[x=B, y expr=\thisrow{DFAR_mean_005}+(3*\thisrow{DFAR_var_005}), col sep=comma] {\datagraphtwo} ;

\addplot [name path=unitwo_lower, draw=none, fill opacity=0.3, forget plot] table[x=B, y expr=\thisrow{DFAR_mean_005}-(3*\thisrow{DFAR_var_005}), col sep=comma] {\datagraphtwo} ;

\addplot [\unicolor!30, forget plot, fill opacity = 0.5] fill between[of=unitwo_upper and unitwo_lower];

\addplot [name path=unitree_upper, draw=none, fill opacity=0.3, forget plot] table[x=B, y expr=\thisrow{DFAR_mean_01}+(3*\thisrow{DFAR_var_01}), col sep=comma] {\datagraphtwo} ;

\addplot [name path=unitree_lower, draw=none, fill opacity=0.3, forget plot] table[x=B, y expr=\thisrow{DFAR_mean_01}-(3*\thisrow{DFAR_var_01}), col sep=comma] {\datagraphtwo} ;

\addplot [\unitree!30, forget plot, fill opacity = 0.5] fill between[of=unitree_upper and unitree_lower];

\addplot [name path=unifour_upper, draw=none, fill opacity=0.3, forget plot] table[x=B, y expr=\thisrow{DFAR_mean_02}+(3*\thisrow{DFAR_var_02}), col sep=comma] {\datagraphtwo} ;

\addplot [name path=unifour_lower, draw=none, fill opacity=0.3, forget plot] table[x=B, y expr=\thisrow{DFAR_mean_02}-(3*\thisrow{DFAR_var_02}), col sep=comma] {\datagraphtwo} ;

\addplot [\unifour!30, forget plot, fill opacity = 0.5] fill between[of=unifour_lower and unifour_upper];

\end{axis}

\end{tikzpicture}
  \caption{}
  \label{simulation fig2}
\end{subfigure}
\caption{Simulation results.
Figure~\ref{simulation fig1} presents the relationship between Principal's rewards and the approximation factor. Figure~\ref{simulation fig2} describes the connection between Principal's rewards and the budget.}
\end{figure*}
In this section, we provide an empirical demonstration of $\model$ with generated data, focusing on finite-horizon DDPs for simplicity. 
\paragraph{Generating MDPs} Ideally, we would conduct the experimental validation on real-world data. However, the rewards of Agent and Principal are application-specific and intricate to obtain; thus, we would have to synthesize them. Consequently, we decided to focus on simple instances and a straightforward generating process.

We construct an MDP characterized by a layer graph, featuring an initial state and consisting of five layers. Within each layer, we include $10$ states. There are $10$ actions with deterministic transitions from each state, each leading to a different state in the next layer. We pick these medium-size scales to allow for non-efficient computation of the optimal bonus scheme (see below). The rewards of Agent and Principal associated with each action are randomly and uniformly sampled from $[0, 1]$. The rewards are independent both between Agent and Principal and among different actions. 
\paragraph{Algorithms} We employ $\MDPPFr$ and a \emph{brute force} algorithm to compute the optimal value for the Principal.\footnote{The code for the simulations is available at \href{https://github.com/boaztait/principal_agent_MDP}{https://github.com/boaztait/principal\_agent\_MDP}} The brute force algorithm starts the calculation from the initial state and recursively visits each of its child states, exhaustively iterating every trajectory of the underlying MDP.
\paragraph{Experiment Setup} We report the mean results of 10,000 generated MDP instances. We used a standard PC with intel Core i7-9700k CPU and 16GB RAM for running the simulations. The entire execution (for the 10,000 runs) took roughly 15 hours.

\subsection{Results}

Figure~\ref{simulation fig1} illustrates the relationship between the Principal's utility and the discretization factor $\varepsilon$. We set $B=1$ and compute $V^P$ as a function of $\varepsilon$, resulting in three distinct curves:
\begin{itemize}
\item The green (solid) curve represents the optimal principal's utility $V^P_\star$ obtained through the brute force algorithm.
\item The orange (dashed) curve corresponds to the formal guarantee from Corollary~\ref{corollary:alg_deterministic_dag_UP_MDP_approximation} for $B' = B-H\varepsilon$, where $H=5$ is the horizon.
\item The blue (dotted) curve denotes the utility of $\MDPPFr$.
\end{itemize}
Each curve represents the mean of 10,000 generated instances, with error bars set at three standard deviations—though these are almost negligible given the high number of generated samples. As anticipated, the results of $\MDPPFr$ for every $\varepsilon$ surpass the lower bound from Corollary~\ref{corollary:alg_deterministic_dag_UP_MDP_approximation} and fall below the optimal Principal's utility.

Notably, the $\MDPPFr$ curve exhibits non-linearity, particularly pronounced for high values of $\varepsilon$. This non-linearity stems from the intricacies of the term $\varepsilon \lfloor \nicefrac{1}{\varepsilon} \rfloor$. For clarity, consider an MDP with a single layer, where the Agent selects one of 10 actions from the initial state and then deterministically transitions to a terminal state. The discretized rewards are ${ n \cdot \varepsilon }_{n=0}^{\lfloor \nicefrac{1}{\varepsilon} \rfloor}$, and when $B=1$, the Principal can implement any policy. However, since $\MDPPFr$ considers only discretized rewards, the highest attainable reward for the Principal is limited to $\varepsilon \lfloor \nicefrac{1}{\varepsilon} \rfloor$.

Furthermore, if the highest reward falls within the $[\varepsilon \lfloor \nicefrac{1}{\varepsilon} \rfloor ,1]$ interval, the Principal can only achieve $\varepsilon \lfloor \nicefrac{1}{\varepsilon} \rfloor$. However, conditioning on this event, the optimal reward becomes $V^P = \varepsilon \lfloor \nicefrac{1}{\varepsilon} \rfloor + \frac{1 - \varepsilon \lfloor \nicefrac{1}{\varepsilon} \rfloor}{2}$. The discrepancy $\frac{1 - \varepsilon \lfloor \nicefrac{1}{\varepsilon} \rfloor}{2}$ between $\MDPPFr$ and the optimal reward introduces non-smoothness in the $\MDPPFr$ curve.

Figure~\ref{simulation fig2} explores the relationship between the Principal's budget and utility. We use $\MDPPFr$ to compute $V^P$ for values of $B$ in the interval $[0, 2]$ with steps of $0.1$ and display four curves corresponding to $\varepsilon \in \{0.01, 0.05, 0.1, 0.2\}$. Additionally, we plot a fifth green (solid) curve representing Principal's optimal utility $V^P_\star$ obtained by the brute force algorithm for each $B$ value. This curve could also be thought of as the result of $\MDPPFr$ with $\varepsilon \rightarrow 0$.

Initially, note that with a budget of $B=0$, all curves yield $V^P=2.5$. This arises from the fact that Agent's optimal policy consists of five transitions, and his rewards are independent of Principal rewards, leading to Principal's expected reward being the expected sum of five uniform $[0,1]$ random variables. Moreover, the curves exhibit strict dominance: Lower values of $\varepsilon$ result in higher $V^P$ that converge to $V^P_\star$.

Finally, the approximation of $V^P$ becomes more accurate for higher values of $B$ for any level of approximation $\varepsilon$. This is due to Corollary~\ref{corollary:alg_deterministic_dag_UP_MDP_approximation}, which guarantees that $\MDPPFr$ returns an $H\varepsilon$ approximation with respect to the instance $\tilde{I} = (S, A, P, \tilde{R}^A, \tilde{R}^P, H, B + H\varepsilon)$, where in our case $H=5$. As $B$ increases, the proportion between the actual budget $B$ and the budget of $\tilde I$ (which is $B + H\varepsilon$) increases to 1, rendering the effect of the additional $H\varepsilon$ negligible.

\section{Discussion}

This paper introduces a novel approach to principal-agent modeling over MDPs, where the principal has a limited budget to shape the agent's reward.  We propose two efficient algorithms designed for two broad problem classes: Stochastic trees and finite-horizon DDPs. We experimentally validate some of our theoretical findings via simulations in {\ifnum\Includeappendix=1{Appendix~\ref{sec:simulations}.}\else{the appendix.}\fi} Further, we also study several extensions that we relegate to {\ifnum\Includeappendix=1{Appendix~\ref{appn: extensions}}\else{the appendix}\fi} in the interest of space. Future work can include better algorithms for our classes of instances as well as more general classes of instances. Another direction for future work is addressing learning scenarios, i.e., when the principal, agent or both have incomplete information.

\section*{Acknowledgements}
This project has received funding from the European Research Council (ERC) under the European Union’s Horizon 2020 research and innovation program (grant agreement No. 882396), by the Israel Science Foundation, the Yandex Initiative for Machine Learning at Tel Aviv University and a grant from the Tel Aviv University Center for AI and Data Science (TAD).

\bibliographystyle{abbrvnat}
\bibliography{ms}

\begin{thebibliography}{46}
\providecommand{\natexlab}[1]{#1}
\providecommand{\url}[1]{\texttt{#1}}
\expandafter\ifx\csname urlstyle\endcsname\relax
  \providecommand{\doi}[1]{doi: #1}\else
  \providecommand{\doi}{doi: \begingroup \urlstyle{rm}\Url}\fi

\bibitem[Altman(1999)]{cmdp}
E.~Altman.
\newblock \emph{Constrained Markov decision processes}, volume~7.
\newblock CRC press, 1999.

\bibitem[Arora and Doshi(2021)]{survey_irl}
S.~Arora and P.~Doshi.
\newblock A survey of inverse reinforcement learning: Challenges, methods and progress.
\newblock \emph{Artificial Intelligence}, 297:\penalty0 103500, 2021.

\bibitem[Banihashem et~al.(2022)Banihashem, Singla, Gan, and Radanovic]{banihashem2022admissible}
K.~Banihashem, A.~Singla, J.~Gan, and G.~Radanovic.
\newblock Admissible policy teaching through reward design.
\newblock In \emph{Proceedings of the AAAI Conference on Artificial Intelligence}, volume~36, pages 6037--6045, 2022.

\bibitem[Ba{\c{s}}ar and Olsder(1998)]{bacsar1998dynamic}
T.~Ba{\c{s}}ar and G.~J. Olsder.
\newblock \emph{Dynamic noncooperative game theory}.
\newblock SIAM, 1998.

\bibitem[Battaglini(2005)]{battaglini2005long}
M.~Battaglini.
\newblock Long-term contracting with markovian consumers.
\newblock \emph{American Economic Review}, 95\penalty0 (3):\penalty0 637--658, 2005.

\bibitem[Bolton and Dewatripont(2004)]{contract_theory}
P.~Bolton and M.~Dewatripont.
\newblock \emph{Contract theory}.
\newblock MIT press, 2004.

\bibitem[Castro(2020)]{castro2020scalable}
P.~S. Castro.
\newblock Scalable methods for computing state similarity in deterministic markov decision processes.
\newblock In \emph{Proceedings of the AAAI Conference on Artificial Intelligence}, volume~34, pages 10069--10076, 2020.

\bibitem[Chakraborty et~al.(2023)Chakraborty, Bedi, Koppel, Manocha, Wang, Huang, and Wang]{chakraborty2023aligning}
S.~Chakraborty, A.~S. Bedi, A.~Koppel, D.~Manocha, H.~Wang, F.~Huang, and M.~Wang.
\newblock Aligning agent policy with externalities: Reward design via bilevel rl.
\newblock \emph{arXiv preprint arXiv:2308.02585}, 2023.

\bibitem[Chen et~al.(2022)Chen, Yang, Li, Wang, Yang, and Wang]{chen2022adaptive}
S.~Chen, D.~Yang, J.~Li, S.~Wang, Z.~Yang, and Z.~Wang.
\newblock Adaptive model design for markov decision process.
\newblock In \emph{International Conference on Machine Learning}, pages 3679--3700. PMLR, 2022.

\bibitem[Chen et~al.(2023)Chen, Wu, Wu, and Yang]{chen2023learning}
S.~Chen, J.~Wu, Y.~Wu, and Z.~Yang.
\newblock Learning to incentivize information acquisition: Proper scoring rules meet principal-agent model.
\newblock \emph{arXiv preprint arXiv:2303.08613}, 2023.

\bibitem[Deb(2011)]{deb2011multi}
K.~Deb.
\newblock \emph{Multi-objective optimisation using evolutionary algorithms: an introduction}.
\newblock Springer, 2011.

\bibitem[Devlin and Kudenko(2012)]{dynamic_reward_shaping}
S.~M. Devlin and D.~Kudenko.
\newblock Dynamic potential-based reward shaping.
\newblock In \emph{Proceedings of the 11th International Conference on Autonomous Agents and Multiagent Systems}, pages 433--440. IFAAMAS, 2012.

\bibitem[Dutting et~al.(2021)Dutting, Roughgarden, and Talgam-Cohen]{dutting2021complexity}
P.~Dutting, T.~Roughgarden, and I.~Talgam-Cohen.
\newblock The complexity of contracts.
\newblock \emph{SIAM Journal on Computing}, 50\penalty0 (1):\penalty0 211--254, 2021.

\bibitem[Gan et~al.(2022)Gan, Han, Wu, and Xu]{optimal_coordination_principal-agent_problems}
J.~Gan, M.~Han, J.~Wu, and H.~Xu.
\newblock Optimal coordination in generalized principal-agent problems: A revisit and extensions.
\newblock \emph{arXiv preprint arXiv:2209.01146}, 2022.

\bibitem[Garey and Johnson(1979)]{guide_theory_np_completeness}
M.~R. Garey and D.~S. Johnson.
\newblock \emph{Computers and intractability}, volume 174.
\newblock freeman San Francisco, 1979.

\bibitem[Grzes(2017)]{reward_shaping_epsisodic_rl}
M.~Grzes.
\newblock Reward shaping in episodic reinforcement learning.
\newblock In \emph{Proceedings of the 16th International Conference on Autonomous Agents and Multiagent Systems}, pages 565--573. ACM, 2017.

\bibitem[Hadfield-Menell and Hadfield(2019)]{hadfield2019incomplete}
D.~Hadfield-Menell and G.~K. Hadfield.
\newblock Incomplete contracting and ai alignment.
\newblock In \emph{Proceedings of the 2019 AAAI/ACM Conference on AI, Ethics, and Society}, pages 417--422, 2019.

\bibitem[Hartline and Roughgarden(2008)]{hartline2008optimal}
J.~D. Hartline and T.~Roughgarden.
\newblock Optimal mechanism design and money burning.
\newblock In \emph{Proceedings of the fortieth annual ACM symposium on Theory of computing}, pages 75--84, 2008.

\bibitem[Ho et~al.(2014)Ho, Slivkins, and Vaughan]{ho2014adaptive}
C.-J. Ho, A.~Slivkins, and J.~W. Vaughan.
\newblock Adaptive contract design for crowdsourcing markets: Bandit algorithms for repeated principal-agent problems.
\newblock In \emph{Proceedings of the fifteenth ACM conference on Economics and computation}, pages 359--376, 2014.

\bibitem[Holmstr{\"o}m(1979)]{holmstrom1979moral}
B.~Holmstr{\"o}m.
\newblock Moral hazard and observability.
\newblock \emph{The Bell journal of economics}, pages 74--91, 1979.

\bibitem[Hu et~al.(2020)Hu, Wang, Jia, Wang, Chen, Hao, Wu, and Fan]{hu2020learning}
Y.~Hu, W.~Wang, H.~Jia, Y.~Wang, Y.~Chen, J.~Hao, F.~Wu, and C.~Fan.
\newblock Learning to utilize shaping rewards: A new approach of reward shaping.
\newblock \emph{Advances in Neural Information Processing Systems}, 33:\penalty0 15931--15941, 2020.

\bibitem[Kamenica(2019)]{bayesian_persuasion_information_design}
E.~Kamenica.
\newblock Bayesian persuasion and information design.
\newblock \emph{Annual Review of Economics}, 11:\penalty0 249--272, 2019.

\bibitem[Laffont(2003)]{laffont2003principal}
J.-J. Laffont.
\newblock \emph{The principal agent model}.
\newblock Edward Elgar Publishing, 2003.

\bibitem[Mannor et~al.(2022)Mannor, Mansour, and Tamar]{MannorMT-RLbook}
S.~Mannor, Y.~Mansour, and A.~Tamar.
\newblock \emph{Reinforcement Learning: Foundations}.
\newblock Online manuscript; \url{https://sites.google.com/view/rlfoundations/home}, 2022.
\newblock accessed March-05-2023.

\bibitem[Martello and Toth(1990)]{martello1990knapsack}
S.~Martello and P.~Toth.
\newblock \emph{Knapsack problems: algorithms and computer implementations}.
\newblock John Wiley \& Sons, Inc., 1990.

\bibitem[Ng et~al.(1999)Ng, Harada, and Russell]{policy_invariance_reward_transformation}
A.~Y. Ng, D.~Harada, and S.~Russell.
\newblock Policy invariance under reward transformations: Theory and application to reward shaping.
\newblock In \emph{Icml}, volume~99, pages 278--287. Citeseer, 1999.

\bibitem[Post and Ye(2015)]{post2015simplex}
I.~Post and Y.~Ye.
\newblock The simplex method is strongly polynomial for deterministic markov decision processes.
\newblock \emph{Mathematics of Operations Research}, 40\penalty0 (4):\penalty0 859--868, 2015.

\bibitem[Rakhsha et~al.(2020)Rakhsha, Radanovic, Devidze, Zhu, and Singla]{rakhsha2020policy}
A.~Rakhsha, G.~Radanovic, R.~Devidze, X.~Zhu, and A.~Singla.
\newblock Policy teaching via environment poisoning: Training-time adversarial attacks against reinforcement learning.
\newblock In \emph{International Conference on Machine Learning}, pages 7974--7984. PMLR, 2020.

\bibitem[Randl{\o}v and Alstr{\o}m(1998)]{randlov1998learning}
J.~Randl{\o}v and P.~Alstr{\o}m.
\newblock Learning to drive a bicycle using reinforcement learning and shaping.
\newblock In \emph{ICML}, volume~98, pages 463--471, 1998.

\bibitem[Ross(1973)]{principal_agent_problem_book}
S.~A. Ross.
\newblock The economic theory of agency: The principal's problem.
\newblock \emph{The American economic review}, 63\penalty0 (2):\penalty0 134--139, 1973.

\bibitem[Stadie et~al.(2020)Stadie, Zhang, and Ba]{stadie2020learning}
B.~Stadie, L.~Zhang, and J.~Ba.
\newblock Learning intrinsic rewards as a bi-level optimization problem.
\newblock In \emph{Conference on Uncertainty in Artificial Intelligence}, pages 111--120. PMLR, 2020.

\bibitem[Sutton and Barto(2018)]{Sutton2018}
R.~S. Sutton and A.~G. Barto.
\newblock \emph{Reinforcement Learning: An Introduction}.
\newblock The MIT Press, second edition, 2018.

\bibitem[Wang et~al.(2022)Wang, Wang, and Gong]{wang2022bi}
L.~Wang, Z.~Wang, and Q.~Gong.
\newblock Bi-level optimization method for automatic reward shaping of reinforcement learning.
\newblock In \emph{International Conference on Artificial Neural Networks}, pages 382--393. Springer, 2022.

\bibitem[Wiering and Van~Otterlo(2012)]{wiering2012reinforcement}
M.~A. Wiering and M.~Van~Otterlo.
\newblock Reinforcement learning.
\newblock \emph{Adaptation, learning, and optimization}, 12\penalty0 (3):\penalty0 729, 2012.

\bibitem[Wiewiora et~al.(2003)Wiewiora, Cottrell, and Elkan]{methods_advising_rl}
E.~Wiewiora, G.~W. Cottrell, and C.~Elkan.
\newblock Principled methods for advising reinforcement learning agents.
\newblock In \emph{Proceedings of the 20th international conference on machine learning (ICML-03)}, pages 792--799, 2003.

\bibitem[Wu et~al.(2022)Wu, Zhang, Feng, Wang, Yang, Jordan, and Xu]{wu2022markov}
J.~Wu, Z.~Zhang, Z.~Feng, Z.~Wang, Z.~Yang, M.~I. Jordan, and H.~Xu.
\newblock Markov persuasion processes and reinforcement learning.
\newblock In \emph{ACM Conference on Economics and Computation}, 2022.

\bibitem[Xiao et~al.(2019)Xiao, Guo, Jiang, Lv, Chen, Zhu, and Yang]{cmd_budget_allocation}
S.~Xiao, L.~Guo, Z.~Jiang, L.~Lv, Y.~Chen, J.~Zhu, and S.~Yang.
\newblock Model-based constrained mdp for budget allocation in sequential incentive marketing.
\newblock In \emph{Proceedings of the 28th ACM International Conference on Information and Knowledge Management}, pages 971--980, 2019.

\bibitem[Xiao et~al.(2020)Xiao, Wang, Chen, Tang, and Yang]{xiao2020optimal}
S.~Xiao, Z.~Wang, M.~Chen, P.~Tang, and X.~Yang.
\newblock Optimal common contract with heterogeneous agents.
\newblock In \emph{Proceedings of the AAAI Conference on Artificial Intelligence}, volume~34, pages 7309--7316, 2020.

\bibitem[Yu and Ho(2022)]{yu2022environment}
G.~Yu and C.-J. Ho.
\newblock Environment design for biased decision makers.
\newblock In \emph{Proceedings of the International Joint Conference on Artificial Intelligence (IJCAI)}, 2022.

\bibitem[Zhang and Conitzer(2021)]{zhang2021automated}
H.~Zhang and V.~Conitzer.
\newblock Automated dynamic mechanism design.
\newblock \emph{Advances in Neural Information Processing Systems}, 34:\penalty0 27785--27797, 2021.

\bibitem[Zhang and Parkes(2008)]{zhang2008value}
H.~Zhang and D.~C. Parkes.
\newblock Value-based policy teaching with active indirect elicitation.
\newblock In \emph{AAAI}, volume~8, pages 208--214, 2008.

\bibitem[Zhang and Zenios(2008)]{dynamic_principal_agent_hidden_info}
H.~Zhang and S.~Zenios.
\newblock A dynamic principal-agent model with hidden information: Sequential optimality through truthful state revelation.
\newblock \emph{Operations Research}, 56\penalty0 (3):\penalty0 681--696, 2008.

\bibitem[Zhang et~al.(2009)Zhang, Parkes, and Chen]{zhang2009policy}
H.~Zhang, D.~C. Parkes, and Y.~Chen.
\newblock Policy teaching through reward function learning.
\newblock In \emph{Proceedings of the 10th ACM conference on Electronic commerce}, pages 295--304, 2009.

\bibitem[Zhang et~al.(2022{\natexlab{a}})Zhang, Cheng, and Conitzer]{efficient_plan_participate_constraint}
H.~Zhang, Y.~Cheng, and V.~Conitzer.
\newblock Efficient algorithms for planning with participation constraints.
\newblock In \emph{Proceedings of the 23rd ACM Conference on Economics and Computation}, pages 1121--1140, 2022{\natexlab{a}}.

\bibitem[Zhang et~al.(2022{\natexlab{b}})Zhang, Cheng, and Conitzer]{plan_participate_constraint}
H.~Zhang, Y.~Cheng, and V.~Conitzer.
\newblock Planning with participation constraints.
\newblock In \emph{Proceedings of the AAAI Conference on Artificial Intelligence}, volume~36, pages 5260--5267, 2022{\natexlab{b}}.

\bibitem[Zhuang and Hadfield-Menell(2020)]{zhuang2020consequences}
S.~Zhuang and D.~Hadfield-Menell.
\newblock Consequences of misaligned ai.
\newblock \emph{Advances in Neural Information Processing Systems}, 33:\penalty0 15763--15773, 2020.

\end{thebibliography}

{\ifnum\Includeappendix=1{
\appendix
\onecolumn

\section{Statements and Proofs Omitted From Section~\ref{sec:model}} 

\subsection{Non-continuous relationship between Principal's utility and the budget} \label{dis vp and B section}

In this section, we provide an example of the discontinuous relationship between Principal's utility and the budget allocation. Consider the example illustrated in Figure~\ref{dis_example}. The underlying MDP has a deterministic transition function. Agent can choose actions from $\{left, right\}$ leading to $s_l$ and $s_r$, respectively. By choosing action $left$, Agent gets a reward of $1$, and Principal gets $0$. By choosing $right$, Agent receives $0$ and Principal $1$, namely $R^A(s_0, right) = 1$, $R^P(s_0, right) = 0$, $R^A(s_0, left) = 0$ and $R^P(s_0, left) = 1$.

Denote the two available policies $\pi_{left}, \pi_{right}$. Without a bonus reward, Agent plays $\pi_{right}$ and gets a reward of $1$, while Principal gets $0$. To implement $\pi_{left}$, Principal can allocate bonus of $R^B(s_0, left) = 1$. In such a case, Agent is indifferent between the two policies and, according to our assumption, will pick $\pi_{left}$.
Any other bonus allocation $R^B(s_0, left) = 1-\varepsilon$ for $\varepsilon > 0$ will not change Agent's default policy since it would grant Agent a reward of $1-\varepsilon$, which is less than what he gets by playing $\pi_{right}$.

\begin{figure*}
\centering
   \begin{tikzpicture}[scale=0.35]    
        \node (s0) [state] at (0, 0) {$s_0$};
        \node (s1) [state] at (4, -3.5) {$s_r$};
        \node (s2) [state] at (-4, -3.5) {$s_l$};

        \node[red, scale=0.9] at (3.6, -0.1) {$R^A=1$};
        \node[blue, scale=0.9] at (3.6, -1) {$R^P=0$};

        \node[red] at (-3.6, -0.1) {$R^A=0$};
        \node[blue] at (-3.9, -1) {$R^P=1$};
        
        \path[-stealth, thick, auto]
            (s0) edge (s1)
            (s0) edge (s2);

    \end{tikzpicture}
\caption{An illustrative instance to demonstrate the non-continuous relationship between the principal's utility and budget allocation.} \label{dis_example}
\end{figure*}
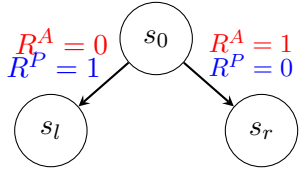

\subsection{Relaxing the Tie-Breaking Assumption} \label{section: more then one policy for RB}
Recall that our analysis assumes that if Agent is indifferent between several policies, he is breaking ties in favor of the policy most beneficial for Principal (Subsection~\ref{subsec: optimization}). To relax this assumption, Principal can add a tiny bonus that will break Agent's indifference. We formalize this in the following Lemma~\ref{lemma: more than one policy rb}. We remark that the proof below assumes an acyclic layout.
\begin{lemma} \label{lemma: more than one policy rb}
Let $(S, A, P, R^A, R^P, H)$ be an MDP with an acyclic layout. Fix any $R^B$, such that $\abs{\mathcal{A}(R^A + R^B)} > 1$, if there exist $\pi_\star \in \mathcal{A}(R^A + R^B)$ such that $V(\pi_\star, R^P) > V(\pi, R^P)$ for any $\pi \in \mathcal{A}(R^A + R^B) \setminus \{ \pi_\star \}$ Then, for any small constant $\varepsilon >0$, there exists $\tilde{R}^B$ such that $\norm{\tilde{R}^B}_1=\norm{R^B}_1+\varepsilon$ and $\pi_\star = \mathcal{A}(R^A + \tilde{R}^B)$.
\end{lemma}
\begin{proofof}{lemma: more than one policy rb}
Let $d(s)$ denote the depth of state $s$. We define a new function $\tilde{R}$ according to

\begin{align*}
& \tilde{R}(s,a)
\begin{cases}
    2^{d(s)}\delta & \mbox{if $\pi^\star(s) = a$}, \\ 
    0 & \mbox{otherwise} 
\end{cases}.
\end{align*}

We show that the new bonus function $\tilde{R}^B = R^B + \tilde{R}$ implements only policy $\pi^\star$.

Assume in contradiction that there exists $\pi \in \mathcal{A}(R^A + \tilde{R}^B) \setminus \{ \pi^\star \}$. Since $\pi \neq \pi^\star$, there exists state $s$ such that $\pi(s) \neq \pi^\star(s)$. For that state $s$ it holds that
\begin{equation}
\begin{aligned}
   V_s(\pi, R^A + \tilde{R}^B) &= V_s(\pi, R^A + R^B + \tilde{R}) \\
   &= V_s(\pi, R^A + R^B) + V_s(\pi, \tilde{R}) \\ &\leq V_s(\pi, R^A + R^B) + (2^{d(s)} - 1)\delta \\
   &\leq V_s(\pi^\star, R^A + R^B) + (2^{d(s)} - 1)\delta \\
   &< V_s(\pi^\star, R^A + R^B) + 2^{d(s)}\delta \\
   &\leq V_s(\pi^\star, R^A + R^B )+V_s(\pi^\star,\tilde{R}) \\
   &= V_s(\pi^\star, R^A + \tilde{R}^B).
\end{aligned}
\end{equation}

The first inequality is a sum over $\tilde{R}$ over all the states forward from state $s$. The second inequality is due to $\pi^\star \in \mathcal{A}(R^A + R^B)$. The last inequality is due to
\[ V_s(\pi^\star, \tilde{R}) = \tilde{R}(s, \pi^\star(s)) + \sum_{s'\in Child(s)}P(s,\pi^\star(s),s')V_{s'}(\pi^\star, \tilde{R}) \geq 2^{d(s)}\delta; 
\]
therefore, policy $\pi^\star = \mathcal{A}(R^A + \tilde{R}^B)$. We conclude this lemma by setting $\varepsilon = 2^{H + 1}$.
\end{proofof}

\subsection{Existence of a Minimal Implementation (Definition~\ref{def: minimal implementation})} \label{appn: model definition}
\begin{lemma} \label{lemma: minimal implementation}
If $\pi$ is B-implementable, then there exists a minimal implementable $R^B$ of it.
\end{lemma}

\begin{proofof}{lemma: minimal implementation}
Fix a B-implementable policy $\pi$. Let $\mathcal{R}_\pi$ denote the set of feasible bonus reward functions that induce it,
\[ \mathcal{R}_\pi = \left\{R^B \mid \sum_{s,a}{R^B(s,a)} \leq B, \pi \in \mathcal{A}(R^A + R^B) \right\}.\]
Since $\pi$ is B-implementable, $\mathcal{R}_\pi$ is non-empty. Let $R\in \mathcal R_\pi$ denote one of its elements. We can use $R$ to redefine $\mathcal{R}_\pi$ as follows: 
\[ \mathcal{R}_\pi = \left\{R^B \mid \sum_{s,a}{R^B(s,a)} \leq B, V(\pi,R^A+R^B) \geq V(\pi, R^A+R)  \right\},\]
where $V(\pi, R^A+R)$ is a constant. Recall that $V$ is a linear function, which means that $\mathcal{R}_\pi$ is a closed set. This means that $B_\pi = \{ \sum_{s,a}{R^B(s,a)} \mid R^B \in \mathcal{R}_\pi\}$ is a closed set that attains its infimum; thus, a minimal implementation must exist.
\end{proofof}


\subsection{NP-Hardness} \label{appendix: np-hard}
\begin{proofof}{thm: UP-MDP NP Hard knapsack}
We show a reduction from the 0/1 Knapsack problem to the $\model$ problem.
We use the same construction as in Example~\ref{example:sketch} from Section~\ref{subsection: warmup example} for the reduction.

The Knapsack problem receives a set of values $\{v_i\}_{i=1}^n \in \mathbb{R}$, costs $\{c_i\}_{i=1}^n \in \mathbb{R}$, and a maximum weight capacity $C \in \mathbb{R}$. The optimization problem is as follows:
\begin{align*}
&\max_{\{x_i\}} \sum_{i=1}^n x_iv_i\\
&\sum_{i=1}^n x_ic_i \leq C \label{eq: knapsack problem} \tag{P2}\\
&x_i \in \{0,1\} \quad \forall i \in \{1,2,..,n\}
\end{align*}
We now describe a polynomial time reduction to the $\model$ problem.
Given a Knapsack instance $\left(\{v_i\}_{i=1}^n,\{c_i\}_{i=1}^n, C \right)$, we construct an instance  $(S, A, P, R^P, R^A, H, B)$ according to Figure~\ref{fig:example2} with $H=2$ and $B=C$. Every item in the Knapsack corresponds to a gadget with three states. The root $s_{0}$ leads to every gadget (or state $s_i$ in the figure) of a total of $n$ gadgets with equal probability of $\frac{1}{n}$. Transitioning from $s_i$ is deterministic: There are two possible actions from state $s_i$, $left$ and $right$, leading to $s_{i,l}$ and $s_{i,r}$, respectively. If Agent picks $right$, the rewards are $R^P(s_i, right)=R^A(s_i, right) = 0$. If Agent picks $left$, the rewards are $R^P(s_i, left) = v_i$, $R^A(s_i, left) = -c_i$. Namely, Agent prefers acting $right$, but Principal prefers he acts $left$. 

Since Knapsack is NP-hard, the next lemma suggests $\model$ is NP-hard as well.
\begin{lemma} \label{lemma:up-mdp knapsack value}
Given a Problem~\eqref{eq: knapsack problem} instance $I$ with $n$ items, let $\tilde I$ be the constructed Problem~\eqref{eq: principles problem} instance. $I$ has a solution with value $V^{\star}$ if and only if $\tilde I$ has a solution with value $\frac{v^{\star}}{n}$.
\end{lemma}
The proof of Lemma~\ref{lemma:up-mdp knapsack value} appears after the proof of Theorem~\ref{thm: UP-MDP NP Hard knapsack}. According to Lemma~\ref{lemma:up-mdp knapsack value}, given any instance of Problem~\eqref{eq: knapsack problem}, we can solve Problem~\eqref{eq: principles problem} using our construction and get a solution to Problem~\eqref{eq: knapsack problem}. This concludes the proof of the theorem.
\end{proofof}

\begin{proofof}{lemma:up-mdp knapsack value}
Let $\pi$ denote an arbitrary policy in $\mathcal{A}(R^A + R^B)$.
We begin with an auxiliary observation. Notice that in every state $s_i$, $\forall i \in [n]$, the agent is better off choosing $left$ if and only if
\begin{equation*}
R^A(s_i, left) + R^B(s_i, left) \geq R^A(s_i, right) \Longleftrightarrow -c_i + R^B(s_i, left) \geq 0,
\end{equation*}
or, equivalently, if $R^B(s_i, left) \geq c_i$. Therefore, the utility $V(\pi, R^P)$ is

\begin{align} \label{eq: knapsack UP-MDP value}
 & V(\pi, R^P) = \frac{1}{n}\sum_{i=1}^n I_{R^B(s_i, left) \geq c_i}v_i.
\end{align}

We are ready to prove the lemma. In the first direction, assume that the Knapsack problem has an optimal solution $\{x_i\}$ with a value of $v^\star=\sum_{i=1}^n x_iv_i$. We shall use $\{x_i\}$ to construct a candidate solution $R^B$ for the $\model$ problem as follows:
\begin{align} \label{eq: knapsack RB}
& R^B(s_i, left) = 
\begin{cases}
    c_i & \mbox{if $x_i = 1$}, \\ 
    0 & \mbox{otherwise} 
\end{cases}.
\end{align}
Notice that our construction of $R^B$ is feasible, since
\[ 
\sum_{s,a}R^B(s,a) = \sum_{i=1}^n x_i R^B(s_i, left) \leq C = B,
\]
where the second-to-last inequality follows from $(x_i)_i$ being a feasible Knapsack solution. In addition, playing $R^B$ yields a reward of $\frac{1}{n} v^\star$. To see this, we combine Equations~\eqref{eq: knapsack RB} and~\eqref{eq: knapsack UP-MDP value} and get

\begin{align*}
& V(\pi,R^P) = \frac{1}{n}\sum_{i=1}^n I_{x_i = 1}v_i= \frac{1}{n}\sum_{i=1}^n x_iv_i =  \frac{1}{n} v^\star.
\end{align*}

This completes the first direction of the proof. 

In the other direction, assume that $R^B$ is a $\model$ solution with value $V(\pi, R^P) = \frac{1}{n}v^\star$. Define the sequence $(x_i)_i$ such that
\begin{align*}
& x_i = 
\begin{cases}
    1 & \mbox{if $R^B(s_i, left) \geq c_i$}, \\ 
    0 & \mbox{otherwise} 
\end{cases}.
\end{align*}
Notice that $(x_i)_i$ is a feasible Knapsack solution, since
\[ 
\sum_{i=1}^n x_ic_i \leq \sum_{i=1}^n R^B(s_i, left) \leq \sum_{s,a} R^B(s, a) \leq B = C,
\]
where the second-to-last inequality follows from $R^B$ being a feasible $\model$ solution. Finally, using $\{x_i\}$ as a solution for the Knapsack problem yields

\[ 
\sum_{i=1}^n x_iv_i = \sum_{i=1}^n I_{R^B(s_i, left) \geq c_i} v_i = n V(\pi, R^P),
\]

where the last equality is due to Equation~\eqref{eq: knapsack UP-MDP value}.
\end{proofof}

\section{Statements and Proofs Omitted From Section~\ref{sec: stochastic trees}}

\subsection{Proof of Observation~\ref{lemma: same agent reward}}
Fix a policy $\pi$. We prove the observation by induction over the states. The base case is all leaf states. This holds due to similar arguments to those given in Lemma~\ref{lemma: delta q minimal implementation}; hence, it is omitted. 

Next, fix a state $s$ and assume the argument holds for every $s' \in Child(s)$. Due to that,
\begin{align*}
V_s(\pi, R^A + R^B) = R^A(s,\pi(s)) + R^B(s,\pi(s))+\sum_{s'} P(s,\pi(s),s') V_{s'}(\pi, R^A + R^B)
\end{align*}
Using the inductive assumption on $V_{s'}(\pi, R^A + R^B)$,
\begin{align}\label{eq:ttyty}
V_s(\pi, R^A + R^B) = R^A(s,\pi(s)) + R^B(s,\pi(s))+\sum_{s'} P(s,\pi(s),s') V_{s'}(\pi^A, R^A + R^B).
\end{align}
Recall that Equation~\eqref{eq:diff for minimal} implies
\begin{align}\label{asdasdasvgv}
&R^B(s, \pi(s)) = Q^{\pi^A}(s, \pi^A(s), R^A) - Q^{\pi^A}(s, \pi(s), R^A) \nonumber \\
&= R^A(s, \pi^A(s)) - R^A(s, \pi(s)) \\
&\quad +\sum_{s'} P(s,\pi^A(s),s') V_{s'}(\pi^A, R^A + R^B)-\sum_{s'} P(s,\pi(s),s') V_{s'}(\pi^A, R^A + R^B) \nonumber
\end{align}
Combining Equations\eqref{eq:ttyty} and \eqref{asdasdasvgv}, we get
\begin{align*}
V_s(\pi, R^A + R^B) =  R^A(s,\pi(s)) + \sum_{s'} P(s,\pi^A(s),s') V_{s'}(\pi^A, R^A + R^B) = V_s(\pi, R^A).
\end{align*}
This concludes the proof of the observation.

\subsection{OCBA Algorithm Implementable For $k$ Children} \label{apn subsection: budget to k children}

\begin{algorithm}
\begin{algorithmic}[1]
\caption{Optimal Children Budget Allocation ($\OCBA$)} \label{alg: dynamic programming optimal allocation}
\REQUIRE $s, a, b, U^P$ 
\ENSURE $U^P_a(s,b)$

\STATE $\forall i \in \{1,2,...,|Child(s)|\}$, $b' \in \mathcal{B}$ $M(i, b') \gets 0$

\FOR{$i \in \{1, 2, ... ,|Child(s)|\}$}
    \FOR{$b' \in \mathcal{B}, b' \leq b$}
        \STATE $M(i, b') \gets \underset{\substack{\tilde{b} \in \mathcal{B} \\ \tilde{b} \leq b'}}{\max}\{M(i-1, b' - \tilde{b}) + p(s, a, s_i)U^P(s_i, \tilde{b})\}$ \label{OA Line: Update M}
    \ENDFOR
\ENDFOR

\STATE $U^P_a(s,b) \gets M(|Child(s)|, b)$
\RETURN $U^P_a(s,b)$ 

\end{algorithmic}
\end{algorithm}
\begin{proposition}\label{prop:OCBA}
Fix an input $(s,a,b,U^P)$ for Algorithm~\ref{alg: dynamic programming optimal allocation}, and let $\textbf{X} \in \mathcal{B}^{\abs{Child(s)}}$ such that $\sum_{i=1}^{ \abs{Child(s)}} \textbf{X}_i = b$. The output of the Algorithm~\ref{alg: dynamic programming optimal allocation} satisfies 
\[U^P_a(s,b) = \max_{\textbf X}{\sum_{i=1}^{ \abs{Child(s)}} P(s,a,s_i)U^P(s_i, X_i)},
\]
where $s_i \in Child(s)$.
\end{proposition}
\begin{proofof}{prop:OCBA}
Algorithm~\ref{alg: dynamic programming optimal allocation} constructs the table $M$ iteratively using $U^P$. This is a standard construction in dynamic programming, see~\citet{martello1990knapsack}.
\end{proofof}

\subsection{Extracting $R^B$ In Algorithm~\ref{alg: general stochastic tree UP-MDP}}\label{subsec:extract}

\begin{algorithm}
\begin{algorithmic}[1]
\caption{Extract $R^B$ from Line~\ref{STUM line: extract RB from TK} of Algorithm~\ref{alg: general stochastic tree UP-MDP}} \label{alg: Extract stochastic RB}
\REQUIRE $U^P, U^P_a$
\ENSURE $R^B$

\STATE Init $R^B(s, a) \gets 0$ for all $s\in S$, $a \in A$
\STATE Init $E(s) \gets B$ for all $s \in S$ \label{alg extract E}
\STATE $curr \gets \{ s_{0} \}$

\WHILE{$curr \neq \emptyset$}
    \STATE pop $s \gets curr$ with the lowest depth
    \STATE for all $a \in A(s)$, $r(a) \gets Q^{\pi_A}(s, \pi^A(s), R^A) - Q^{\pi_A}(s, a, R^A)$ \label{line: extract RB update ra}

    \STATE $a_{opt} \gets \underset{a \in A(s)}{\argmax}\{ U^P_a(s, E(s) - \underset{\substack{r \in \mathcal{B} \\ r \leq r(a)}}{\max}\{r\})\}$

    \STATE $R^B(s, a_{opt}) \gets r(a_{opt})$

    \STATE for all $s' \in Child(s)$, set $E(s')$ the allocation of $\OCBA(s, a_{opt}, E(s)-\underset{\substack{r \in \mathcal{B} \\ r \leq r(a_{opt})}}{\max}\{r\})$ \label{extract_RB line: get b distribution}

    \STATE $curr \gets curr \cup \{s' | E(s') > 0, s' \in Child(s) \}$
    
\ENDWHILE

\RETURN $R^B$

\end{algorithmic}
\end{algorithm}

In this subsection, we present a dynamic programming approach to implement Line~\ref{STUM line: extract RB from TK} in  Algorithm~\ref{alg: general stochastic tree UP-MDP}. We run a top-down computation to allocate the budget $R^B(s,a)$ according to the optimal utilities $U^P, U^P_a$, which were already computed by Algorithm~\ref{alg: general stochastic tree UP-MDP}. For completeness, we present this process explicitly.

We start with a brief explanation. In Line~\ref{alg extract E} , we initialize $E(s)$ for every $s\in S$, which is an upper bound on the budget we allocate in the sub-tree of $s$. Once we know how much budget we place on each state, we can direct that budget to the relevant action. Line~\ref{extract_RB line: get b distribution} calculates the budget allocation for each child of state $s$.

\begin{lemma} \label{lemma: Extract stochastic RB}
Let $I = (S, A, P, R^A, R^P, H, B)$ be a $k$-ary tree, and let $V^P_\star$ be the optimal solution for $I$. let $\tilde{I}$ be the identical instance but with a budget $B + \varepsilon \abs{S}$ for a small constant $\varepsilon > 0$. Let $U^P$ and $U^P_a$ be Principal's utilities from Algorithm~\ref{alg: general stochastic tree UP-MDP}, Then executing Algorithm~\ref{alg: Extract stochastic RB} takes a run time of $O(\abs{S}(k(\nicefrac{B}{\varepsilon}) + \abs{A}))$, and for its output $R^B$ there exists $\pi \in \mathcal{A}(R^A + R^B)$ such that $V(\pi, R^P) \geq V^P_\star$.
\end{lemma}

\begin{proofof}{lemma: Extract stochastic RB}
Lemma~\ref{lemma: Extract stochastic RB} iterates in the opposite order of Algorithm~\ref{alg: PARETO_FRONTIER} and uses the same statements. We prove each statement after the proof of Theorem~\ref{thm: general stochastic tree UP-MDP} and hence it is omitted.
\end{proofof}

\subsection{Guarantees of Algorithm~\ref{alg: general stochastic tree UP-MDP}}

\begin{proofof}{thm: general stochastic tree UP-MDP}
Fix an instance $I$. We use Lemma~\ref{lemma: delta q minimal implementation} below to justify the assignment in Line~\ref{STUM line: update r of a} in Algorithm~\ref{alg: general stochastic tree UP-MDP}.
\begin{lemma} \label{lemma: delta q minimal implementation}
Fix a state $s'\in S$, and let $a\in A$ be any action in an acyclic MDP. Further, denote the policy $\pi_{(s,a')}$ such that for any $s \in S$,
\begin{align*}
& \pi_{(s',a)}(s) = 
\begin{cases}
\pi^A(s) & s \neq s' \\
a & \textnormal{otherwise}
\end{cases}.
\end{align*}
If $\pi_{(s,a')}$ is B-implementable, then the following $R^B$ is a minimal implementation of $\pi$:
\begin{align*}
& R^B_{(s',a)}(s,a) = 
\begin{cases}
    V_s(\pi^A, R^A) - Q^{\pi^A}(s, a, R^A) & \mbox{if $s = s'$}, \\ 
    0 & \mbox{otherwise} 
\end{cases}.
\end{align*}
\end{lemma}
Leveraging Lemma~\ref{lemma: delta q minimal implementation}, we show that
\begin{corollary} \label{cor: delta q minimal implementation}
Fix a B-implementable policy $\pi$. The bonus function $R^B$ such 
\[
R^B = \sum_{(s,a): s\in S, a \in A, \pi(s)=a} R^B_{(s,a)}
\]
is a minimal implementation of policy $\pi$.
\end{corollary}
The proof of Lemma~\ref{lemma: delta q minimal implementation} and Corollary~\ref{cor: delta q minimal implementation} appear at the end of this proof. Next, denote by $V^P_{\star}(s, b)$ the optimal utility when starting from $s$ with a budget $b$, where we let $V^P_{\star}(s, b) = V^P_{\star}(s, 0)$ in case $b<0$. Further, let $\Sub(s)$ denote the number of states in the subtree of any state $s\in S$, not including the root. For instance, $\alpha(s)=0$ for every leaf.
\begin{lemma} \label{lemma: Up correctness}
Let $I = (S, A, P, R^A, R^P, H)$ be a $k$-ary tree, fix any state $s\in S$, a budget unit $b \in \mathcal{B}$, and let $\varepsilon > 0$ be some small constant. Then, the computed $U^P(s,b)$ in Line~\ref{STUM line: update T of action} in Algorithm~\ref{alg: general stochastic tree UP-MDP} satisfies $U^P(s,b) \geq V^P_{\star}(s, b-\varepsilon \cdot\Sub(s))$.
\end{lemma}

According to Lemma~\ref{lemma: Up correctness}, $U^P(s_0, B+\varepsilon \abs{S}) \geq V^{P}_{s_0,\star}(B) = V^P_\star$. From Lemma~\ref{lemma: Extract stochastic RB}, Algorithm~\ref{alg: Extract stochastic RB} construct the bonus function $\tilde{R}^B$ that induces $U^P(s_0, B + \varepsilon \abs{S})$. This concludes the approximation guarantees.

To calculate the runtime, we first start with the loop of the backward induction.
For every state $s$, the runtime of one iteration of the while loop in Line~\ref{STUM: curr iteration} is the sum of runtimes of Lines~\ref{STUM: pop from curr}-\ref{STUM: curr add parent}. Line~\ref{STUM: pop from curr} can be implemented using a stack and therefore takes $O(1)$. Line~\ref{STUM line: update r of a} runs for every $a \in A(s)$ and therefore runs in $O(\abs{A})$. Line~\ref{STUM line: update T of probability} is executed $O(\abs{A}\abs{\mathcal{B}})$ times, and the runtime of each execution is $O(k \abs{\mathcal{B}}^2)$, thus the total runtime of Line~\ref{STUM line: update T of probability} is $O(k \abs{A} \abs{\mathcal{B}}^3)$. Line~\ref{STUM line: update T of action} runs  $O(\abs{A}\abs{\mathcal{B}})$ times, and lastly Line~\ref{STUM: curr add parent} can be implemented in $O(1)$ from the same argument as Line~\ref{STUM: pop from curr}. Therefore, the total runtime of each iteration in the backward induction is given by $O(k \abs{A} \abs{\mathcal{B}}^3)$.
The while loop is executed once per state. Thus, the overall runtime of the backward induction is $O(\abs{S} k \abs{A} \abs{\mathcal{B}}^3)$. Finally, Line~\ref{STUM line: extract RB from TK} is given by Lemma~\ref{lemma: Extract stochastic RB} and runs in $O(\abs{S}(k\abs{\mathcal{B}} + \abs{A}))$. To conclude, the total runtime of Algorithm~\ref{alg: general stochastic tree UP-MDP} is $O(\abs{S} k \abs{A} \abs{\mathcal{B}}^3)$. By substituting $\abs{\mathcal{B}} = \nicefrac{B}{\varepsilon}$ we get $O(\abs{S} \abs{A} k (\nicefrac{B}{\varepsilon})^3)$.
\end{proofof}

\begin{proofof}{lemma: delta q minimal implementation}
We develop Agent's utility with the given $R^B$, 
\begin{equation*}
    \begin{aligned}
        V_{s'}(\pi_{(s', a)}, R^A + R^B) &= V_{s'}(\pi_{(s', a)}, R^A) + V_{s'}(\pi_{(s', a)}, R^B) \\
        &= Q^{\pi_{(s', a)}}({s'}, \pi_{(s', a)}(s), R^A) + V_{s'}(\pi^A, R^A) - Q^{\pi^A}({s'}, a, R^A) \\
        &= V_{s'}(\pi^A, R^A) + \sum_{s \in S}P(s',a,s)V_s(\pi_{(s', a)}, R^A) - \sum_{s \in S}P(s',a,s)V_s(\pi^A, R^A) \\
        &= V_{s'}(\pi^A, R^A) + \sum_{s \in S}P(s',a,s)V_s(\pi^A, R^A) - \sum_{s \in S}P(s',a,s)V_s(\pi^A, R^A) \\
        &= V_{s'}(\pi^A, R^A),
    \end{aligned}
\end{equation*}
where the second and third equalities are due to the definition of the $Q$ function. The second to last equality is due to $\pi(s) = \pi^A(s)$ for every $s \neq s'$.

Since $R^B(s,a) = 0$ for every $s \neq s'$ and any $a$, it holds that $V_s(\pi, R^A + R^B) = V_s(\pi^A, R^A)$. Overall, we conclude that $\pi \in \mathcal{A}(R^A + R^B)$; hence,  Definition~\ref{def: minimal implementation} suggests that $R^B$ is a minimal implementation of policy $\pi$.
\end{proofof}

\begin{proofof}{cor: delta q minimal implementation}
Assume by contradiction that there exists another $\tilde{R}^B$ such that $\sum_{s,a}\tilde{R}^B(s,a) \leq \sum_{s,a} R^B(s,a)$, and for every state
\begin{equation} \label{eq: min ratio}
 V_s(\pi, R^A + \tilde{R}^B) = V_s(\pi, R^A + R^B).
\end{equation}

The utility of each state is given by
\begin{equation} \label{eq: explicit value}
V_s(\pi, R^A + R^B) = R^B(s, \pi(s)) + \sum_{s'}P(s,\pi(s), s')V_{s'}(\pi, R^A + R^B)    
\end{equation}

Summing over the states in Equation~\eqref{eq: explicit value}
\begin{align*}
&\sum_{s}{\tilde{R}^B(s, \pi(s))} + \sum_{s}{ \sum_{s'}P(s,a,s')V_{s'}(\pi, R^A + \tilde{R}^B) } \\
&= \sum_{s}{\tilde{R}^B(s, \pi(s))} + \sum_{s}{ \sum_{s'}P(s,a,s')V_{s'}(\pi, R^A + R^B) } \\
&< \sum_{s}{R^B(s, \pi(s))} + \sum_{s}{ \sum_{s'}P(s,a,s')V_{s'}(\pi, R^A + R^B) },
\end{align*}

But this is in contradiction to Equation~\eqref{eq: min ratio}.

We now show that for every minimum implementation and for every state, it holds that
$ V_s(\pi, R^A + R^B) = V_s(\pi^A, R^A + R^B) $.

Denote $F(R) = \sum_s{V_s(\pi, R)}$. We use the following proposition to prove our corollary.

\begin{proposition} \label{prop: f min rb}
Let $R_1$ such that $F(R_1) = C_1$, $\sum_{s,a}R_1(s,a) = \min \sum_{s,a}R(s,a)$, and let $R_2$ such that $F(R_2) = C_2$, $\sum_{s,a}R_2(s,a) = \min \sum_{s,a}R(s,a)$, then if $C_1 > C_2$ then $\sum_{s,a}R_1(s,a) > \sum_{s,a}R_2(s,a)$.
\end{proposition}

Notice that when taking $R^B$ in Proposition~\ref{prop: f min rb}, it must satisfy $V_s(\pi, R^A + R^B) \geq \max_{\pi'} V_s(\pi', R^A + R^B)$ for every state to be Agent's best response. In this case, the proposition holds because both $F$ and the constraints are linear functions.

Assume by contradiction that there exists $\tilde{R}^B$ such that $\sum_{s,a}\tilde{R}^B(s,a) < \sum_{s,a}R^B(s,a)$, and 
$V_s(\pi, R^A + \tilde{R}^B) \geq V_s(\pi, R^A + R^B)$ for every state, where for at least one state it holds $V_s(\pi, R^A + \tilde{R}^B) > V_s(\pi, R^A + R^B)$.

Expanding the expression of the utility
$ V_s(\pi, R^A + R^B) = V_s(\pi, R^A) + V_s(\pi, R^B) $, and
summing over all the states results in
\begin{align*}
\sum_s{V_s(\pi, R^A)} + \sum_s{V_s(\pi, R^B)} &= \sum_s{V_s(\pi, R^A + R^B)} \\
&< \sum_s{V_s(\pi, R^A + \tilde{R}^B)} \\
&= \sum_s{V_s(\pi, R^A)} + \sum_s{V_s(\pi, \tilde{R}^B)}.
\end{align*}

We got $F(R^B) < F(\tilde{R}^B)$, therefore $\sum_{s,a}R^B(s,a) < \sum_{s,a}\tilde{R}^B(s,a)$, in contradiction to our assumption.
\end{proofof}

\begin{proofof}{lemma: Up correctness}
Fix an instance $I$ and $\varepsilon >0$. We want to show that 
\begin{equation}\label{eq: condition for Up}
U^P(s,b) \geq V^P_{\star}(s, b-\varepsilon \Sub({s}))
\end{equation}
holds for every pair $(s,b)$ with $s\in S$ and $b \in \mathcal B$. The above holds trivially for leaves with any non-positive budget units. 

Next, fix an arbitrary state $s$ and a budget unit $b$ that are not covered by these degenerate cases. Due to the correctness of $\OCBA$ and Line~\ref{STUM line: update T of action}, we know that $U^P(s,b)$ is Principal's maximal reward if she can only assign $\varepsilon$-discrete budget to each state-action pair. Let $\beta(s')$ be the budget assigned to $(s',\pi(s'))$ for the induced optimal policy of Agent.

Additionally, let $\beta^\star(s')$ be \emph{the optimal}  budget assigned to an action $(s,a)$ for some action $a$ under $V^P_\star(s, b-\varepsilon \alpha(s))$. Notice that
\[
=\sum_{s'}\beta^\star(s') \leq \sum_{s'}\varepsilon\ceil{ \frac{\beta^\star(s')}{\varepsilon}  } \leq b- \varepsilon \alpha(s)  + \varepsilon \alpha(s)= b;
\]
thus, the assignment $\beta'(s')=\varepsilon\ceil{ \frac{\beta^\star(s')}{\varepsilon}  }$, which uses a budget of at most $b$, is a candidate budget assignment for $U^P(s,b)$. Due to monotonicity, Agent's policy induced by $\{\beta'(s')\}$ is inferior to $\{\beta(s')\}$. This suggests Equation~\eqref{eq: condition for Up} holds.
\end{proofof}
\section{Statements and Proofs Omitted From Section~\ref{sec: deterministic DAG}}

\subsection{$\model$ With DDP Layout Is Hard} \label{sec: deterministic hard}
\begin{theorem}\label{thm: deterministic dag UP-MDP NP-Hard}
The class of instances of $\model$ with a DDP layout is NP-Hard.
\end{theorem}

\begin{proofof}{thm: deterministic dag UP-MDP NP-Hard}

To prove Theorem~\ref{thm: deterministic dag UP-MDP NP-Hard} we introduce two more problems that we will use later on. First, we use Observation~\ref{observation: deterministic and pareto} to rewrite Problem~\ref{eq: principles problem} for the class of DDP instances and name it the Constrained Optimal Policy Problem ($COPP$). Given $(S, A, P, R^A, R^P, H, B)$ the goal is to find policy $\pi$ that maximizes Principal's utility and satisfy $V(\pi, R^A) > V(\pi^A, R^A) - B$. In DDPs each policy $\pi$ induces a path $\tau$, and therefore we define the $COPP$ problem using paths instead of policies. Formally
\begin{align*}
& \qquad \max_{\tau}{V(\tau, R^P)} \label{eq: COOP}  \tag{P3} \\
& \qquad V(\tau, R^A) \geq V(\tau^A, R^A) - B,
\end{align*}
where $\tau^A$ is the path induced by policy $\pi^A$.
\begin{observation}\label{lemma:copp solution UP-MDP}
Fix an instance $(S, A, P, R^P, R^A, H, B)$.  Then, $V^P_\star$ is an optimal solution with policy $\pi$ for Problem~\ref{eq: principles problem} if and only if $V^P_\star$ is an optimal solution with policy $\pi$ for Problem~\ref{eq: COOP}.
\end{observation}

The proof of Observation~\ref{lemma:copp solution UP-MDP} follows directly from Observation~\ref{observation: deterministic and pareto} as the two problems share the same set of solutions.

To prove Theorem~\ref{thm: deterministic dag UP-MDP NP-Hard}, we show a reduction from Weight-Constrained Shortest Path ($WCSPP$) \cite{guide_theory_np_completeness} to Problem~\ref{eq: COOP}. Given weighted graph G=(V, E) with weights $\{w_e\}_{e \in E}$, costs $\{c_e\}_{e \in E}$, and maximum weight $W \in \mathbb{R}$, where the weights and costs are defined on the edges, the goal is to find the least-cost path while keeping the total weights below a certain threshold. Formally,
\begin{align*}
& \min_{\tau}{\sum_{e \in \tau}c_e} \label{eq: WCSSP} \tag{P4}\\
&  \sum_{e \in \tau}w_e \leq W.
\end{align*}
We use the following statements to prove our theorem. The proof of all the auxiliary statements is given at the end of this proof.
\begin{proposition}[For instance, \cite{guide_theory_np_completeness}]
Problem~\ref{eq: WCSSP} is NP-Hard.
\end{proposition}

\begin{lemma}\label{lemma:copp is np}
There exists a polynomial-time reduction from Problem~\ref{eq: WCSSP} to Problem~\ref{eq: COOP}.
\end{lemma}

To finish the proof of Theorem~\ref{thm: deterministic dag UP-MDP NP-Hard}, we will use a reduction from the $COPP$ problem. From Observation~\ref{lemma:copp solution UP-MDP} the two problems share the same solutions and therefore it concludes our proof.
\end{proofof}


\begin{proofof}{lemma:copp is np}
We show that that Problem~\ref{eq: COOP} is equivalent to Problem~\ref{eq: WCSSP} by setting $c_e = -R^P(s,a)$, $w_e =-R^A(s,a)$ and $W = -(V(\tau^A, R^A)-B)$ where edge $e \in E$ in graph $G$ corresponds to the state-action pair $s,a$ in $S$ and $A$, therefore he translation between Problem~\ref{eq: WCSSP} to Problem~\ref{eq: COOP} is done in $O(\abs{E})$. Under those relations, we see that 
\begin{align*}
& V(\tau, R^P) = \sum_{s,a\in \tau}R^P(s,a) = -\sum_{e\in \tau}c_e \\
& V(\tau, R^A) = \sum_{s,a\in \tau}R^A(s,a) = -\sum_{e\in \tau}w_e
\end{align*}
therefore we get
\begin{align}
& \nonumber \min_\tau \sum_{e\in \tau}c_e = \min_{\tau}{(-V(\tau, R^P))} = \max_{\tau}{V(\tau, R^P)} \\
& -V(\tau, R^A) = \sum_{e\in \tau}w(e) \leq W = -(V(\tau^A, R^A) -B) \label{eq: wcssp weight to Va}
\end{align} 
The result of rearranging Equation~\eqref{eq: wcssp weight to Va} is $V(\tau, R^A) \geq V(\tau^A, R^A) - B$.
Thus given any instance of Problem~\ref{eq: WCSSP}, we can translate the instance and get a solution to Problem~\ref{eq: COOP}, which can then be translated back to a solution of Problem~\ref{eq: WCSSP}. This concludes the proof of the theorem.

We note that we use the negative sign to switch between the minimum in the WCSSP to the maximum in COPP and to flip the inequality condition between the two problems. In the same manner, we could use a variant of the WCSSP where we need to maximize the cost and have the sum of weights above a certain threshold to get the same reduction with positive signs.
\end{proofof}


\subsection{Proof of Observation~\ref{observation: deterministic and pareto}}
In this subsection, we prove Observation~\ref{observation: deterministic and pareto}.

In the first direction, we assume that the policy $\pi$ is B-implementable. Therefore, according to Definition~\ref{def:implementable_policy}, there exists $R^B$ such that
$\pi \in \mathcal{A}(R^A + R^B)$ and $\sum_{s,a} R^B(s,a) \leq B$. Since $\pi \in \mathcal{A}(R^A + R^B)$ then it must satisfy
\[
V(\pi, R^A + R^B) \geq V(\pi', R^A + R^B) 
\]
where $\pi'$ is any other policy. In particular, it is also true for $\pi' = \pi^A$, therefore
\[
V(\pi, R^A + R^B) \geq V(\pi^A, R^A + R^B) = V(\pi^A, R^A) + V(\pi^A, R^B) \geq V(\pi^A, R^A).
\]
Equivalently, 
\[ V(\pi, R^A + R^B)= V(\pi, R^A) + V(\pi, R^B) \geq V(\pi^A, R^A). \]
Rearranging the above,
\[ 
V(\pi, R^A) \geq V(\pi^A, R^A) - V(\pi, R^B) \geq V(\pi^A, R^A) - B,
\]
where the second inequality sign is due to $\sum_{s,a} R^B(s,a) \leq B$.

On the other direction, we assume that there exists $\pi$ such that $V(\pi, R^A) \geq V(\pi^A, R^A) - B$. We construct $R^B$ using the following rule:
\begin{align*}
& R^B(s,a) = 
\begin{cases}
    Q^{\pi^A}(s, \pi^A(s), R^A) - Q^{\pi^A}(s, a, R^A) & \mbox{if $(s,a) \in \tau$}, \\ 
    0 & \mbox{otherwise} 
\end{cases}.
\end{align*}
From Corollary~\ref{cor: delta q minimal implementation}, the constructed $R^B$ is a minimal implementation of policy $\pi$. and from Definition~\ref{def: minimal implementation} it satisfies $\pi \in \mathcal{A}(R^A + R^B)$ and $\sum_{s,a} R^B(s,a) \leq B$.

This completes the proof of Observation~\ref{observation: deterministic and pareto}.

\subsection{Removing Pareto Inefficient Vectors} \label{appendix pareto remove complexity}
The next proposition shows that Line~\ref{call_remove_dominated} of Algorithm~\ref{alg: PARETO_FRONTIER} can be computed in $O(\abs{U}log(\abs{U})$.
\begin{proposition} \label{pareto remove complexity}
Let $U$ be a set of $n$ vectors in $\mathbb R^2$. We can compute 
$Pareto(U)$ in $O(n log(n))$ runtime.
\end{proposition}
\begin{proofof}{pareto remove complexity}
This is a standard procedure (see, e.g., \citet{deb2011multi}); hence the proof is omitted.
\end{proofof}

\subsection{DDP Algorithm Theorem}
\begin{proofof}{thm:alg_deterministic_dag_UP_MDP_approximation}
Let
\[
\mathcal U_s =  \left\{   \left( V_s(\pi, R^A), V_s(\pi, R^P) \right) \in \mathbb R^2\mid \pi \textnormal{ is a policy in the underlying MDP}
\right\}.
\]
\begin{lemma} \label{lemma:vs_pareto_frontier}
Let the input for Algorithm~\ref{alg: PARETO_FRONTIER} be $I=(S, A, P, R^A, R^P, H)$, $I$ is an acyclic and deterministic instance. For every $s\in S$, $U(s)$ in Line~\ref{call_remove_dominated} of Algorithm~\ref{alg: PARETO_FRONTIER} satisfies $U(S)=Pareto(\mathcal U_s)$.
\end{lemma}
Due to Lemma~\ref{lemma:vs_pareto_frontier},
\begin{corollary} \label{corollary: optimal vec on the pareto frontier}
Let $\pi$ be a policy such that $V(\pi, R^P) = \max_{\pi'} V(\pi', R^P)$ and $V(\pi, R^A) \geq V(\pi^A, R^A) - B$. Further, denote $u^\pi=(V(\pi, R^A), V(\pi, R^P))$. Then $U(s_0)$ computed by Algorithm~\ref{alg: PARETO_FRONTIER} contains $u^\pi$.
\end{corollary}
The extraction of the optimal policy in Line~\ref{MDPPF: get RB of policy}  is omitted since it is similar to Algorithm~\ref{alg: general stochastic tree UP-MDP}. 
Finally, the construction of $R^B$ is done according to Corollary~\ref{cor: delta q minimal implementation}.

Now we move to analyze the runtime complexity of Algorithm~\ref{alg: PARETO_FRONTIER}. We start with the loop of the backward induction. Fix a state $s$, the runtime of one iteration of the while loop in Line~\ref{backward_propagate} is the sum of runtimes of Lines~\ref{calc_frontier_for_curr}-\ref{update_curr}. The runtime of Line~\ref{calc_frontier_for_curr} and Line~\ref{update_curr} depends on the backward induction implementation. We can assume the runtime is $O(1)$ by assuming it is implemented using a stack. Line~\ref{val_calculation} depends on the size of the Pareto frontier. According to the discretization, $\abs{U(s)} = 2\frac{H}{\varepsilon}$. Line~\ref{val_calculation} runs over $U(s')$ for every child state $s'$ reachable by taking action $a \in A(s)$, therefore the run time of Line~\ref{val_calculation} is $O(\frac{\abs{A}H}{\varepsilon})$. Afterwards, Line~\ref{call_remove_dominated} runs in $O(\frac{\abs{A}H}{\varepsilon}\log(\frac{\abs{A}H}{\varepsilon}))$ according to Proposition~\ref{pareto remove complexity}. The backward induction loop is executed for every state, and therefore the total run time of the backward induction is $O(\frac{\abs{S}\abs{A}H}{\varepsilon}log(\frac{\abs{A}H}{\varepsilon}))$.

In Line~\ref{deterministic_DAG_search_optimal_val}, finding the optimal utility vector is done in $O(\frac{H}{\varepsilon})$, and the extraction of the policy is in $O(\frac{\abs{A}H^2}{\varepsilon})$. Lastly, Line~\ref{MDPPF: get RB of policy} runs in $O(H)$. To summarize, Algorithm~\ref{alg: PARETO_FRONTIER} runs in $O(\frac{\abs{S}\abs{A}H}{\varepsilon}\log(\frac{\abs{A}H}{\varepsilon}))$.
\end{proofof}

\begin{proofof}{corollary:alg_deterministic_dag_UP_MDP_approximation}
The proof of this corollary follows from the proof of Theorem~\ref{thm:alg_deterministic_dag_UP_MDP_approximation}. Due to the $\varepsilon$-discretization, Principal loses a facor of at most $\varepsilon$ at each state on the chosen path. Similarly, Principal allocates a redundant budget unit of at most $\varepsilon$ at each state on the chosen path. This leads to a utility of least $V^P_\star - H\varepsilon$ with a budget of most $B + H\varepsilon$.
\end{proofof}

\begin{proofof}{lemma:vs_pareto_frontier}
We prove this lemma by induction, where we follow the order of Algorithm~\ref{alg: PARETO_FRONTIER}. The base case is when $s\in Terminal(S)$. Since $s$ is a terminal state, it has no child states, and therefore $U(s)=\{(0, 0)\}$. Additionally $\mathcal U_s = \{(0,0)\}=Pareto(\mathcal U_s)$ as well.

Next, fix an arbitrary state $s$ and assume the inductive assumption holds for every $s' \in Child(s)$. We need to show that  $u \in U(s)$ if and only if $u\in Pareto(\mathcal{U}_s)$.

Fix a utility vector $u \in U(s)$. Line~\ref{val_calculation} suggests that there exists $s'\in Child(s)$ and  $u' \in U(s')$ such that $u' = u - (R^A(s,a), R^P(s,a))$ for some action $a$. If $u \in \mathcal{U}_s$, we are done.

Otherwise, assume by contradiction that $u \notin \mathcal{U}_s$. Due to the inductive assumption, we know that $u' \in \mathcal{U}_{s'}$. Notice that $u = u' + (R^A(s,a), R^P(s,a))$; thus, the only way $\mathcal{U}_s$ does not contain $u$ is if another $\tilde u \in \mathcal{U}_s$ dominates $u$. In such a case, there exists $s'' \in Child(s)$ such that $\tilde u = \tilde u' + (R^A(s,a'), R^P(s,a'))$ for some action $a'$ and some $\tilde u' \in \mathcal{U}_{s'}$. But due to the inductive assumption, $\tilde u' \in U(s')$, and hence $\tilde u \in U(s)$. Since  $\tilde u$ dominates $u$, $u \notin U(s)$ and we reached a 
contradiction. 

The other direction is identical and is therefore omitted.
\end{proofof}

\subsection{Finite Horizon DDP To an Acyclic Layer Hraph} \label{section: ddp to acyclic layered}
In this subsection, we formally describe the construction of an acyclic layer graph using a DDP with horizon $H$.

Let $S, A, P$ be the state space, action space, and transition function of the DDP. Let $s_1,s_2,\dots,s_{\abs{S}}$ denote the states in $S$ for any arbitrary order. Further, let $(S', A, P')$ be the new MDP such that $|S'| = |S|H$. the state space $S'$ is divided into $H$ layers, where in each layer, there are $|S|$ states corresponding to the states of the original DDP. Let $s'_{i,j} \in S'$ be the $i$'th state in the $j$'th layer. The transition function $P'$ is defined as $P'(s'_{i,j}, a, s'_{k,j+1}) = P(s_i, a, s_k)$ $\forall i,k \in |S|$, $j \in \{1,2,\dots H\}$, $\forall a \in A$. Each transition between states in the DDP is a transition between two consecutive layers.
\section{Extensions} \label{appn: extensions}
\subsection{Approximate Agent}
In this subsection, we consider an extension with an \emph{approximate Agent} extension. We assume Agent is willing to forgo an additive $\delta$ fraction of his utility to benefit Principal. Formally, we define the set of $\delta$-approximate policies $\mathcal{A}^\delta$, i.e.,
\[
\mathcal{A}^\delta (R^A+R^B) = \left\{\pi \mid V(\pi,R^A+R^B)  \geq \max_{\pi^\star} V(\pi^\star,R^A+R^B) -\delta   \right\}.
\]
The Principal's problem becomes:
\begin{align}
& \nonumber \max_{R^B}{V(\pi, R^P)} \\
&  \sum_{s\in S, a\in A}{R^B(s, a)} \leq B \label{eq: approximate principles problem}  \tag{$\delta$-P1}\\ 
& \nonumber  {R^B(s, a) \geq 0} \textnormal{ for every } s\in S, a\in A(s) \\
& \nonumber  \color{red}{\pi \in \mathcal{A^\delta}(R^A+R^B)}
\end{align}
Note that we highlighted in red the part that is different than Problem~\ref{eq: principles problem}. 
Consider a modified version of Algorithm~\ref{alg: general stochastic tree UP-MDP}, which we term $\STUMr_\delta$, that only modifies the extraction of $R^B$ in Line~\ref{STUM line: extract RB from TK}. Recall that Subsection~\ref{subsec:extract} with the extraction, and introduces  Algorithm~\ref{alg: Extract stochastic RB} for this task. Consequently, we modify Line~\ref{line: extract RB update ra} to fit Problem~\ref{eq: approximate principles problem}. The modified version is:

\begin{quote}
    for all $a \in A(s)$, 
    $r(a) \gets {\color{red}{\underset{\substack{r \in \mathcal{B} \\ r \leq V_s(\pi^A, R^A) - Q^{\pi_A}(s, a, R^A)}}{\max}\{r\} }}$,
\end{quote}
where again we colored the modified part. Next, we present the guarantees of $\STUMr_\delta$ to the classes of instance we study in Sections~\ref{sec: stochastic trees} and \ref{sec: deterministic DAG}.
\begin{theorem}\label{thm:trees thm extension}
Let $I = (S, A, P, R^A, R^P, H, B)$ be a $k$-ary tree, and let $V^P_\star$ be the optimal solution for $I$ of Problem~\ref{eq: principles problem}. Further, fix  $\delta > 0$ and  let ${R}^{B}$ denote the output of  $\STUMr_\delta(I)$ with $\varepsilon = \nicefrac{\delta}{\abs{S}}$. Then, executing $\STUMr_\delta(I)$ takes a  run time of $O\left(|A||S|k(\nicefrac{B |S|}{\delta})^3\right)$, and its output $R^{B}$ satisfies $V(\pi^{\delta}, R^P) = V^P_\star$ 
for any $\pi^\delta \in \mathcal{A}^{\delta}(R^A + R^{B})$.
\end{theorem}
The proof of Theorem~\ref{thm:trees thm extension} is similar to the proof of Theorem~\ref{thm: general stochastic tree UP-MDP} and is hence omitted.

We obtain results with the same flavor could also be obtained for finite-horizon DDPs, where we change the assignment of $R^B$ during the extraction. Let $\MDPPFr_\delta$ denote the modified algorithm, then
\begin{theorem} \label{thm:ddp thm extension}
Let $I=(S, A, P, R^A, R^P, H, B)$ be an acyclic and deterministic instance, and  let $V^P_\star$ be the optimal solution for $I$ of Problem~\ref{eq: principles problem}. Let $\tilde{I} = (S, A, P, \tilde{R}^A, \tilde{R}^P, H, B)$ be an instance with $\varepsilon$-discrete versions of $R^A$ and $R^P$, $\tilde{R}^A$ and $\tilde{R}^P$, respectively, for a small constant $\varepsilon>0$. Let $R^B$ denote the output of $\MDPPFr_\delta(I)$ for $\delta = H\varepsilon$.

Then, executing $\MDPPFr_\delta(I)$ takes a run time of $O\left(\frac{|S||A|H}{\varepsilon}\log(\frac{|A|H}{\varepsilon})\right)$, and its output $R^B$ satisfies $V(\pi, R^P) \geq  V^P_\star - H\varepsilon$  for any $\pi^\delta \in \mathcal{A}^\delta(R^A + R^B)$.
\end{theorem}
The proof of Theorem~\ref{thm:ddp thm extension} is the same as the proof of Theorem~\ref{thm:alg_deterministic_dag_UP_MDP_approximation} and hence omitted.

\subsection{Unlimited Principal}
In this subsection, we discuss a scenario in which Principal has an unlimited budget. In such a case, Principal can induce any policy she wants. Formally, let $\pi^P$ be Principal's optimal policy, i.e., $\pi^P \in \mathcal A(R^P)$. Principal's goal is to implement her utility-maximizing policy with the lowest possible budget, namely,

\begin{align}
& \nonumber \min{\sum_{s\in S, a\in A}{R^B(s, a)}} \\
&  {R^B(s, a) \geq 0} \textnormal{ for every } s\in S, a\in A(s) \label{eq: principles problem unlimited}  \tag{P5} \\
& \nonumber \pi^P \in \mathcal{A}(R^A+R^B) 
\end{align}
Next, we describe how to solve Problem~\ref{eq: principles problem unlimited} for the classes of instances we analyzed in the paper.
\begin{theorem} \label{thm: unlimited principal}
The optimal solution to Problem~\ref{eq: principles problem unlimited} is the reward function $R^B$ such that
\begin{align*}
& R^B(s,a) = 
\begin{cases}
    Q^{\pi^A}(s, \pi^A(s), R^A) - Q^{\pi^A}(s, a, R^A) & \mbox{if $\pi^P(s) = a$} \\ 
    0 & \mbox{otherwise} 
\end{cases}.
\end{align*}
\end{theorem}
The proof of Theorem~\ref{thm: unlimited principal} follows immediately from Corollary~\ref{cor: delta q minimal implementation}.

\subsection{Principal Cost For State-Action Pair} \label{Principal cost for state-action pair}
Recall that our model assumes that costs are \emph{uniform}: Placing one unit of budget costs the same for every (state, action) pair. In this section, we extend the model to accommodate non-uniform costs. This is the case in real-world situations where, e.g., incentivizing an employee to take on a challenging project might be more costly than motivating them to complete a routine task.

Formally, let $C:\mathbb R_+ \times S \times A \rightarrow \mathbb R_+$ be the cost function, where $C(r,s,a)$ is the cost of placing a reward of $r$ on the (state, action) pair $(s,a)$. The model in Section~\ref{sec:model} is thus a special case where $C(r,s,a)=r$ for every $r\in \mathbb R_+$ and $(s,a)\in S\times A$. The total cost Principal can exhaust is upper bounded by the budget $B$. Consequently, to address non-uniform costs, we can modify Problem~\ref{eq: principles problem} as follows (red color highlights the different part):
\begin{align}
& \nonumber \max_{R^B}{V(\pi, R^P)} \\
&  \sum_{s\in S, a\in A}{\color{red}{C(R^B(s,a), s, a)}} \leq B \label{eq: cost principles problem}  \tag{PC1}\\ 
& \nonumber  {R^B(s, a) \geq 0} \textnormal{ for every } s\in S, a\in A(s) \\
& \nonumber  \pi \in \mathcal{A}(R^A+R^B)
\end{align}
Further, for this variant, we also modify the definition of $B$-implementable policy.
\begin{definition}[$B$-implementable policy]\label{def:implementable_policy costs}
A policy $\pi$ is $B$-implementable if there exists bonus function $R^B$ such that $\sum_{s,a}{C(R^B(s,a), s, a)}\leq B$ and $\pi \in  \mathcal{A}(R^A+R^B)$.
\end{definition}

Modifications of our algorithms obtain similar guarantees to those mentioned before. In the $\STUM$ algorithm, Line~\ref{STUM line: update T of action} would change to
\begin{quote}
    \centering
    \ref{STUM line: update T of action}: for every $b \in \mathcal{B}$, set $U^P(s,b) \leftarrow \max_{a\in A(s)} \{U^P_a(s,b-\underset{\substack{c \in \mathcal{B} \\ c \leq C(r(a), s,a)}}{\max}{c})\}$.
\end{quote}
The proof of the modified Theorem~\ref{thm: general stochastic tree UP-MDP} is identical. 

Dealing with $\MDPPF$ is a bit more tricky. Notice that Observation~\ref{observation: deterministic and pareto} does not hold in its current form: A policy $\pi$ might be $B$-implementable but violate $V(\pi, R^A) \geq V(\pi_A, R^A) - B$. Consider the following modification of Observation~\ref{observation: deterministic and pareto}.
\begin{observation}\label{observation: deterministic and pareto}
If the transition function is deterministic, then a policy $\pi$ is $B$-implementable if and only if 
\[
V(\pi, R^A) \geq V(\pi_A, R^A) - \left[\sum_{(s,a)\in \pi} Q^{\pi^A}(s, \pi^A(s), R^A) - Q^{\pi^A}(s, a, R^A)\right].
\]
\end{observation}
From here on, we describe in high-level detail how to modify $\MDPPF$. First, we use the same discretization for the cost function $C$. Second, instead of storing the Pareto frontier of the rewards, we store triplets: both rewards and the cost. Third, we modify Line~\ref{deterministic_DAG_search_optimal_val}:
\begin{quote}
\centering
    \ref{deterministic_DAG_search_optimal_val}: let $\pi$ such that $u^\pi \leftarrow \underset{\substack{u^\pi \in U(s_0) \\ C(u^\pi) \leq B}}{\argmax} V(\pi, R^P)$, 
\end{quote}
where $C(u^\pi)$ is the entry of the triplet $u^\pi$ that stores the required cost. The modified algorithm has guarantees with a similar flavor to Theorem~\ref{thm:alg_deterministic_dag_UP_MDP_approximation}	and Corollary~\ref{corollary:alg_deterministic_dag_UP_MDP_approximation}.

}{\fi}

\end{document}